\documentclass{article}

\IfFileExists{neurips_2026.sty}{
  \usepackage[preprint]{neurips_2026}
}{
  \usepackage[margin=1in]{geometry}
  \usepackage[numbers]{natbib}
}

\usepackage[T1]{fontenc}
\usepackage[utf8]{inputenc}
\usepackage{microtype}
\usepackage{amsmath}
\usepackage{amssymb}
\usepackage{booktabs}
\usepackage{multirow}
\usepackage{tabularx}
\usepackage{array}
\usepackage{graphicx}
\usepackage{xcolor}
\usepackage{enumitem}
\usepackage{hyperref}
\usepackage{url}
\usepackage{cleveref}
\usepackage{pgfplots}
\pgfplotsset{compat=1.18}
\usepackage{subcaption}
\usepackage{listings}
\lstdefinestyle{pyMin}{
  language=Python,
  basicstyle=\ttfamily\scriptsize,
  keywordstyle=\color{blue!60!black}\bfseries,
  stringstyle=\color{green!50!black},
  commentstyle=\color{gray}\itshape,
  showstringspaces=false,
  columns=fullflexible,
  keepspaces=true,
  belowskip=-2pt, aboveskip=-2pt,
}

\hypersetup{
  colorlinks=true,
  linkcolor=blue!60!black,
  citecolor=blue!60!black,
  urlcolor=blue!60!black
}

\newcolumntype{L}[1]{>{\raggedright\arraybackslash}p{#1}}
\newcommand{\datasetname}{FindStatBench}
\newcommand{\gptmini}{GPT-5.4-mini}

\newcommand{\haiku}{Claude Haiku 4.5}
\newcommand{\sonnet}{Claude Sonnet 4.6}
\newcommand{\geminiflash}{Gemini 2.5 Flash}
\newcommand{\gptosslarge}{gpt-oss-120b}
\newcommand{\gptosssmall}{gpt-oss-20b}
\newcommand{\qwenmoe}{Qwen3-235B-A22B-Instruct}
\newcommand{\qwencoder}{Qwen3-Coder-480B-A35B-Instruct}
\newcommand{\deepseekv}{DeepSeek-V3.1}
\newcommand{\llama}{Llama-3.3-70B-Instruct-Turbo}
\newcommand{\qwensmall}{Qwen2.5-7B-Instruct-Turbo}

\title{\datasetname: Evaluating Large Language Models on Combinatorial Code Synthesis}
\author{Soham Dan\\Scale AI}
\date{}

\begin{document}
\maketitle

\begin{abstract}
We introduce \datasetname{}, an execution benchmark for evaluating large language models on combinatorial code synthesis. Built from \href{https://www.findstat.org/}{FindStat}, the benchmark contains 2{,}329 tasks across 24 collections and 5.52M hidden instances in two executable families: \emph{statistic synthesis} (object\,$\to$\,integer) and \emph{map synthesis} (object\,$\to$\,object). Each task provides a mathematical description and at most five public input--output examples; a model must emit a single Python \texttt{solve} function, with no retrieval, tool use, execution feedback, voting, or reranking. Submissions are scored by exact sandboxed execution on held-out combinatorial objects.

We evaluate eleven systems under this protocol: four closed-source production models and seven open-weight models served through a common inference provider. Three findings motivate \datasetname{} as a stress test for symbolic program induction rather than general software engineering. First, the strongest open- and closed-source systems converge within $\sim$1\,pp instance accuracy, while an oracle over all eleven systems improves task accuracy by only $\sim$10\,pp over the best single model; five-way sampling from one mid-tier model reaches the same ceiling. Second, examples are not uniformly helpful: on several classical named bijections, zero-example prompts produce perfect implementations while five-example prompts collapse to near-zero hidden accuracy and fail even their public examples, suggesting prompt-induced regression away from canonical algorithms. Third, apparent coverage gaps can reflect output-budget mechanics rather than inability: hidden reasoning can consume the visible response budget before code is emitted, a failure mode observed in both closed-source and open-weight reasoning models.

Across all evaluated systems, statistic synthesis remains substantially easier than map synthesis, set partitions and binary trees remain near-zero-accuracy collections, and long prompts induce a sharp accuracy cliff. Cost-normalised, the strongest open-weight systems dominate the closed-source production models on this benchmark. \datasetname{} exposes a distinctive capability gap: current models can often write plausible mathematical code, but exact symbolic rule induction over structured objects remains brittle across scale, specialization, and weight provenance.
\end{abstract}

\section{Introduction}
\label{sec:intro}

Execution-based evaluation has shaped contemporary code generation: HumanEval~\citep{chen2021codex}, MBPP~\citep{austin2021mbpp}, APPS~\citep{hendrycks2021apps}, DS-1000~\citep{lai2023ds1000}, CodeContests~\citep{li2022alphacode}, LiveCodeBench~\citep{jain2024livecodebench}, MultiPL-E~\citep{cassano2022multiple}, and BigCodeBench~\citep{zhuo2024bigcodebench} cover interview problems, library use, contests, and contamination control. Mathematical reasoning suites such as MATH~\citep{hendrycks2021math} and miniF2F~\citep{zheng2021minif2f} probe natural-language reasoning and theorem proving, but say little about a complementary regime: writing short, exact programs from formal mathematical descriptions over combinatorial objects.

\datasetname{} fills that gap. The benchmark is built from FindStat~\citep{findstat}, a public repository of integer-valued statistics and structure-preserving maps on combinatorial objects (permutations, Dyck paths, posets, tableaux, graphs, $\ldots$). For each FindStat page the pipeline emits a closed-book prompt, a sandboxed Python execution harness, and many hidden input--output instances of the same rule. Performance therefore separates ``the model produced code'' from ``the induced rule generalises'' across millions of held-out objects per benchmark run.

\paragraph{Contributions.}
\begin{enumerate}[leftmargin=18pt,topsep=2pt,itemsep=1pt]
  \item A reproducible execution benchmark of 2{,}329 closed-book tasks (1{,}993 statistics + 336 maps) and 5.52M hidden instances spanning 24 collections, and full sandboxed pipeline artifacts.
  \item An eleven-system evaluation covering four closed-source production models and seven open-weight models, spanning various sizes and reasoning efforts.
  \item Dataset-level findings the benchmark exposes: (i) the strongest open and closed models converge to within $\sim$1\,pp on this task and the ensembled frontier is only $\sim$10\,pp above any single model; (ii) on classical named bijections, additional public examples can \emph{regress} accuracy.
  \item Four ablations isolating common phenomena: self-repair recovers +5\,pp on a mid-tier model and +4\,pp on the open-weight frontier model (enough that the repaired open-weight system surpasses the strongest closed-source model on instance accuracy); example-count is non-monotone for maps; pass@5 reaches the same +9.9\,pp ceiling as an 11-model oracle ensemble; and the greedy single-shot evaluation is stable across seeds (std $<$1.3\,pp).
\end{enumerate}

\begin{figure}[t]
\centering
\includegraphics[width=\textwidth]{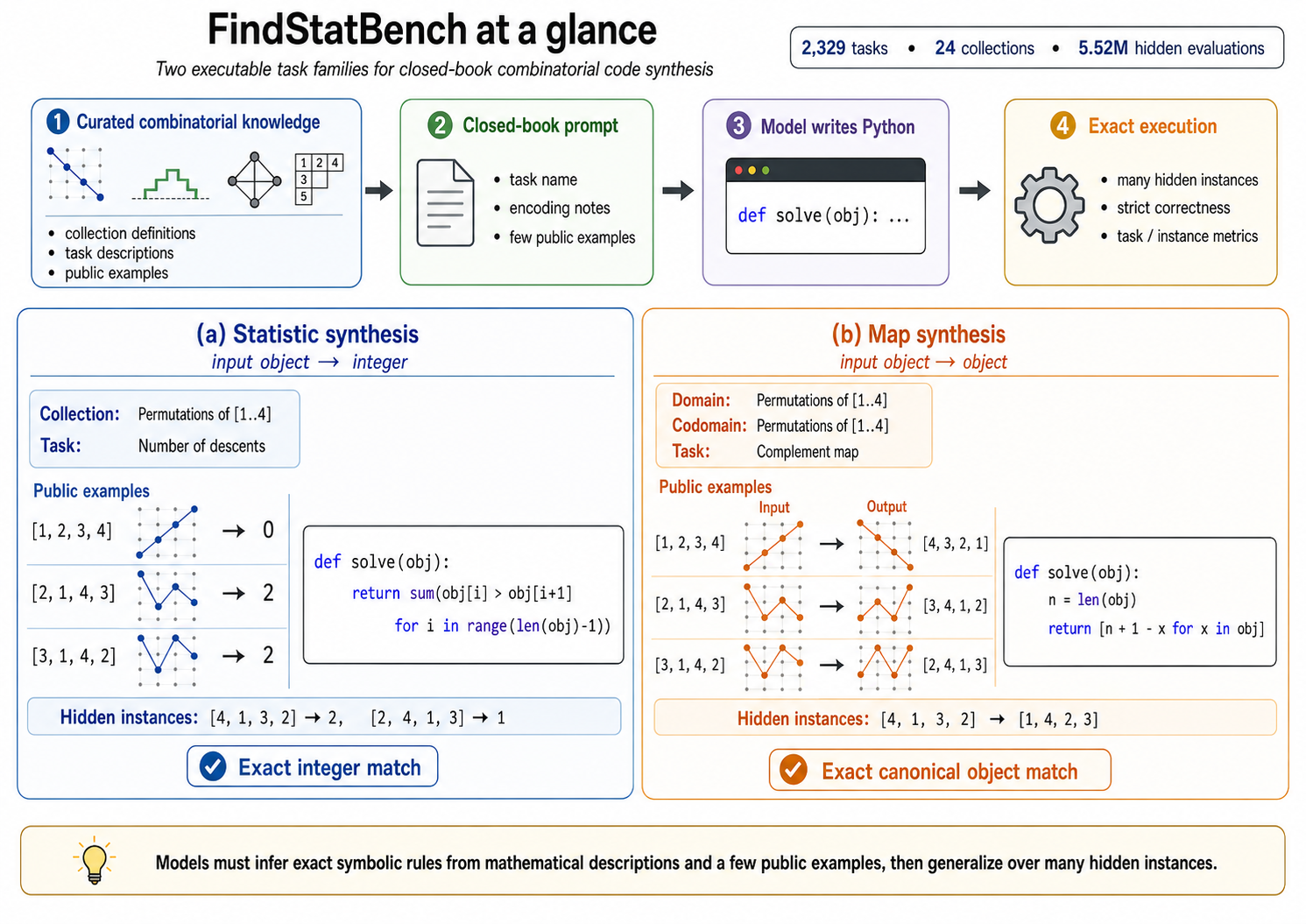}
\caption{\datasetname{} at a glance. Curated FindStat pages become closed-book prompts that require the model to write a Python \texttt{solve(obj)} function inducing an exact symbolic rule; each task is evaluated by exact execution against many hidden instances.}
\label{fig:overview}
\end{figure}

\section{The \datasetname{} Dataset}
\label{sec:dataset}

\paragraph{Source data and crawl.}
\datasetname{} is built from static FindStat pages~\citep{findstat}. A polite crawler (rate-limited, exponentially backed off, with caching and stable on-disk filenames) starts from the homepage and the static catalogue assets (\texttt{/static/all\_stats.json}, \texttt{/static/all\_maps.json}) and parses pages into JSONL artefacts: collections (definitions, encoding notes, canonicalisation hints, enumerative info), statistics (titles, descriptions, references, visible object\,$\to$\,integer pairs), and maps (titles, descriptions, domain/codomain metadata, properties, visible object\,$\to$\,object pairs). Object encodings retain the raw string and add structured parses only when unambiguous. Reproducibility details are in App.~\ref{app:repro}.

\paragraph{Task families.}
\textbf{Statistic synthesis} gives the model a collection name, encoding notes, a textual statistic description, and a few public examples; it must return Python code defining \texttt{solve(obj)} that outputs an integer. \textbf{Map synthesis} adds domain/codomain metadata and any structural properties (e.g.\ injective, graded) and must return the mapped object in the benchmark's canonical encoding. Both families share the same prompt skeleton and execution harness.

\paragraph{Prompt and protocol.}
The benchmark is closed-book in the LLM-evaluation sense: each task is presented in one prompt and the model emits one program, with no retrieval, no tool use, no execution feedback, no majority voting, and no reranking on hidden tests. The released prompt template additionally strips FindStat identifiers, URLs, and explicit source branding to remove the most direct retrieval hooks, but we do not over-claim contamination control: the underlying mathematical descriptions remain public on the web. An open-book vs.\ stripped-identifier ablation on \qwenmoe{} (App.~\ref{app:single-model-ablations}) shows the empirical gap is under 1\,pp on instance accuracy, so the distinction is methodological rather than a strong contamination control on this corpus. The released pipeline supports both prompt variants; we evaluate the stripped-identifier prompt throughout the main paper.

\paragraph{Public/hidden split.}
Tasks are constructed from FindStat's visible image pairs. We require at least four visible pairs (in the cleaned corpus all retained tasks have $\geq$10), choose public examples deterministically by a size-aware key, and use $\sim$30\% of the visible pairs as public examples capped at five; the remainder become hidden instances. The benchmark is therefore \emph{instance-wise}: a task is described once and tested on many held-out objects. The cap-at-five public-example setting yields nearly degenerate prompt support (2{,}325 of 2{,}329 tasks expose five examples, three expose four, one exposes three), which we exploit when probing example-count sensitivity in \cref{sec:examples-derail}.

\begin{table}[t]
\centering
\small
\begin{tabular}{lr}
\toprule
\textbf{Benchmark statistic} & \textbf{Value} \\
\midrule
Collections / statistic tasks / map tasks / total tasks & 24 / 1{,}993 / 336 / 2{,}329 \\
Statistic / map / total hidden instances & 1.93M / 3.59M / 5.52M \\
Visible pairs per task (min / median / max) & 10 / 1{,}200 / 94{,}601 \\
%Tasks dropped after filtering & 0 \\
\bottomrule

\end{tabular}

\caption{Statistics for the FindStat benchmark used in this paper.}
\label{tab:dataset}
\end{table}

\section{Evaluation Protocol}
\label{sec:protocol}

\paragraph{Execution and matching.}
Submissions must define a callable \texttt{solve(obj)}. Statistic tasks use exact integer match; map tasks use exact match after canonicalisation, comparing normalised raw strings and structured forms when available. Code runs in a constrained subprocess sandbox: 2-second timeout per instance, a 256\,MB memory cap when available, a restricted builtin environment, an allowlist of standard-library imports (\texttt{math}, \texttt{itertools}, \texttt{collections}, $\ldots$), and no filesystem or network access. Lightweight AST checks flag syntax errors, disallowed imports, and large literal tables.

\paragraph{Metrics.}
Coverage and correctness separate sharply on this benchmark, so we report several denominators. \textbf{Task accuracy} = perfectly solved tasks $/$ all tasks. \textbf{Instance accuracy} = correct hidden instances $/$ all hidden instances. \textbf{Covered accuracy} conditions on tasks the evaluator accepted as executable (App.~\ref{app:per-model-main}). \textbf{Macro task-instance accuracy} averages per-task hidden-instance accuracy uniformly across tasks.

\paragraph{Models.}
We evaluate four cost-efficient closed-source production models---\gptmini{}, \haiku{}, \sonnet{}, and \geminiflash{}---each via the provider's batch API, and seven open-weight models served via Together~AI's serverless inference: \gptosslarge{} and \gptosssmall{}~\citep{openai_gptoss_2025}; \qwenmoe{} and \qwencoder{}~\citep{qwen3_2025}; \deepseekv{}~\citep{liu2024deepseekv3}; \llama{}~\citep{dubey2024llama3}; and \qwensmall{}~\citep{qwen2_5_2024}. All eleven receive the same closed-book prompt export, generate one program per task in a single turn with no execution-feedback loop, and are scored by the same local sandbox. Provider-specific knobs (reasoning effort, output-token cap, response mode) are reported in App.~\ref{app:model-setup}; the gpt-oss line uses \texttt{max\_tokens=4{,}000} to mitigate the failure mode of \cref{sec:budget-failure}. %Total Together~AI spend across the seven open-weight runs was \$12.96.

\section{Main Results}
\label{sec:main}

\begin{table}[t]
\centering
\small
\resizebox{\columnwidth}{!}{%
\begin{tabular}{lrrrrr}
\toprule
\textbf{Model} & \textbf{Task} & \textbf{Inst.} & \textbf{Stat. inst.} & \textbf{Map inst.} & \textbf{Map / Stat.} \\
\midrule
\multicolumn{6}{l}{\textit{Closed-source production models}} \\
\sonnet{} & 49.29 & 56.04 & 65.82 & 50.79 & 0.77 \\
\gptmini{} & 48.35 & 52.03 & 66.15 & 44.44 & 0.67 \\
\haiku{} & 38.30 & 47.35 & 57.22 & 42.04 & 0.73 \\
\geminiflash{} & 32.33 & 39.51 & 46.97 & 35.50 & 0.76 \\
\midrule
\multicolumn{6}{l}{\textit{Open-weight models (Together AI serverless)}} \\
\gptosslarge{} & 49.25 & 55.13 & 64.02 & 50.34 & 0.79 \\
\gptosssmall{} & 43.84 & 51.27 & 57.89 & 47.71 & 0.82 \\
\qwenmoe{} & 37.78 & 46.43 & 57.18 & 40.65 & 0.71 \\
\qwencoder{} & 34.82 & 39.90 & 53.04 & 32.84 & \textbf{0.62} \\
\deepseekv{} & 31.13 & 39.16 & 48.74 & 34.00 & 0.70 \\
\llama{} & 24.52 & 32.47 & 42.97 & 26.83 & 0.62 \\
\qwensmall{} & 12.75 & 21.15 & 27.02 & 17.99 & 0.67 \\
\bottomrule
\end{tabular}}
\caption{Full-corpus accuracy on \datasetname{} for all eleven evaluated systems (\%, except the final ratio). The map-to-statistic instance-accuracy ratio is below 1 for every model: statistic synthesis is universally easier than map synthesis. The code-specialised \qwencoder{} has the lowest ratio of any frontier-tier model (0.62). App.~\ref{app:per-model-main} adds generation rate, covered, and cost columns; App.~\ref{app:family} adds covered-instance accuracy per family.}
\label{tab:main-results}
\end{table}

\paragraph{The strongest open- and closed-source models converge.}
The strongest open-weight model, \gptosslarge{}, reaches 55.1\% instance accuracy and 49.2\% task accuracy---within 0.92\,pp on instance accuracy of the strongest closed-source model \sonnet{} (56.0\% / 49.3\%) and essentially tied on task accuracy. \gptmini{} (52.0\% / 48.4\%) and \gptosssmall{} (51.3\% / 43.8\%) form the next tier, with the merged \gptmini{} retry slightly ahead on strict instance accuracy but still behind \sonnet{} and \gptosslarge{}. Within the Qwen3 family the general 235B-MoE \qwenmoe{} outperforms the substantially larger code-specialised \qwencoder{} 480B-MoE by +6.5 pp on instance accuracy, and all three open-weight MoE systems beat the dense \llama{} baseline by 7--14\,pp; full per-family detail is in App.~\ref{app:family}.

\paragraph{Coverage and correctness bifurcate, but only for one model.}
Ten of the eleven systems generate executable code on at least 99.3\% of tasks; \gptmini{} is the only model requiring a substantial output-budget retry. Its initial accepted subset was unusually strong (64.2\% covered-task / 68.3\% covered-instance), while the 6{,}400-token retry recovered many previously missing programs with much lower conditional accuracy. \Cref{fig:appendix-gen-covered} visualises this pre-retry coverage--correctness decoupling, and \cref{sec:budget-failure} shows the gap is dominated by a reasoning-budget failure mode shared with the gpt-oss line.\footnote{Per-model partial / zero-pass / syntax-error / generation-miss counts are in App.~\ref{app:failures}.}

\paragraph{Cost rankings invert accuracy rankings.}
\begin{figure}[t]
\centering
\begin{tikzpicture}
\begin{axis}[
  width=0.92\columnwidth, height=5.8cm,
  xlabel={Estimated batch cost (USD)},
  ylabel={Instance accuracy (\%)},
  xmin=-0.5, xmax=24.5, ymin=22, ymax=62,
  xtick={0,5,10,15,20}, ytick={25,30,35,40,45,50,55,60},
  grid=major, grid style={dashed,gray!25},
  tick label style={font=\small}, label style={font=\small},
  legend pos=south east,
  legend style={font=\scriptsize, fill=white, fill opacity=0.85, draw=gray!50},
]
\addplot[only marks, mark=*, mark size=3.2pt, color=blue!70!black] coordinates {
  (21.96, 56.04) (8.13,  52.03) (6.12,  47.35) (1.62,  39.51)
};
\addlegendentry{Closed-source}
\addplot[only marks, mark=triangle*, mark size=3.6pt, color=orange!80!red] coordinates {
  (2.83,  55.13) (1.07,  51.27) (0.88,  46.43) (4.86,  39.90)
  (1.10,  39.16) (1.84,  32.47) (0.37,  21.15)
};
\addlegendentry{Open-weight}
\node[font=\scriptsize, above=2pt] at (axis cs:21.96,56.04) {Sonnet};
\node[font=\scriptsize, above=2pt] at (axis cs:8.13, 52.03) {GPT-mini};
\node[font=\scriptsize, above=2pt] at (axis cs:6.12, 47.35) {Haiku};
\node[font=\scriptsize, right=3pt] at (axis cs:1.62, 39.51) {Gemini};
\node[font=\scriptsize, above=2pt] at (axis cs:2.83, 55.13) {gpt-oss-120b};
\node[font=\scriptsize, above=2pt] at (axis cs:1.07, 51.27) {gpt-oss-20b};
\node[font=\scriptsize, below=2pt] at (axis cs:0.88, 46.43) {Qwen3-235B};
\node[font=\scriptsize, above=2pt] at (axis cs:4.86, 39.90) {Qwen3-Coder-480B};
\node[font=\scriptsize, below=2pt] at (axis cs:1.10, 39.16) {DeepSeek-V3.1};
\node[font=\scriptsize, below=2pt] at (axis cs:1.84, 32.47) {Llama-3.3};
\node[font=\scriptsize, below=2pt] at (axis cs:0.37, 21.15) {Qwen2.5-7B};
\end{axis}
\end{tikzpicture}
\caption{Cost vs.\ instance accuracy for all eleven systems; circles closed-source, triangles open-weight. \gptosslarge{} nearly matches \sonnet{}'s accuracy at one-eighth the cost and strictly dominates the other closed-source providers on this frontier.}
\label{fig:cost-accuracy}
\end{figure}
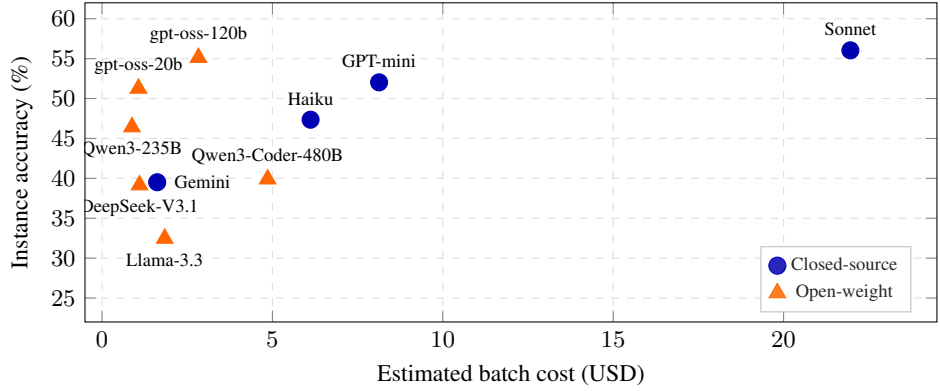
\Cref{fig:cost-accuracy} plots cost against instance accuracy. \sonnet{} delivers the best closed-source accuracy at \$22; \qwenmoe{} reaches 996 perfect tasks per dollar at \$0.88; even \gptosslarge{}, the strongest open-weight model in the paper, costs \$2.83. The benchmark therefore exposes \emph{two distinct leaderboards}: an absolute-accuracy ranking with \gptosslarge{} essentially tied with \sonnet{}, and a cost-efficiency frontier led by open-weight systems at lower budgets. App.~\ref{app:per-model-main} reports the full cost table.

\section{Cross-model phenomena}
\label{sec:phenomena}

We highlight four regularities that hold across all eleven evaluated systems and that the benchmark is structurally well suited to expose.

\paragraph{Statistics are easier than maps for every model.}
On instance accuracy, every system performs better on statistic synthesis than on map synthesis (App.~\ref{app:family}). The gap ranges from $\sim$8\,pp (\geminiflash{}: 47.0\,$\to$\,35.5) to $\sim$20\,pp (\qwencoder{}: 53.2\,$\to$\,32.9). The single outlier within an OSS family is \qwencoder{}, whose map-to-statistic ratio (0.62) is the lowest of any model in the paper, while general-purpose \qwenmoe{} from the same Qwen3 family sits at 0.71. Code specialisation appears to bias toward integer-output tasks specifically rather than helping symbolic synthesis broadly.

\paragraph{Some collections are universally easy, others universally hard.}
\Cref{tab:collection-strip} reports instance accuracy on twelve representative collections for five representative models (the two strongest closed, the two strongest open-weight, and the smallest open-weight). Three patterns emerge that hold for every model in the table. First, the four sequence-/shape-like collections at the top---plane partitions, integer compositions, integer partitions, binary words---are broadly easier than the rest: every model reaches at least 73\% on plane partitions and the strongest systems are competitive on the other three. Second, the two relational collections at the bottom---set partitions and binary trees---are universally near zero: every model is below 6\% on either, the strongest open and closed models are within noise of each other, and even the closed-source-only oracle does not move the floor. These two collections expose deep structural challenges of \datasetname{} that do not yield to scale, code specialisation, MoE architecture, or open/closed provenance. Third, the middle collections---Dyck paths, graphs, posets, lattices, permutations---are differentially difficult by \emph{model}, not just by \emph{collection}: on graphs \gptmini{} reaches 61\% but \qwenmoe{} reaches 51\%, while on Dyck paths \sonnet{} leads at 41\% and \qwenmoe{} falls to 24\%. The benchmark therefore separates \emph{universal} structural difficulty from \emph{model-specific} reasoning quality on a per-collection basis.

\begin{table}[t]
\centering
%\small
\resizebox{\columnwidth}{!}{%
\begin{tabular}{lrrrrrr}
\toprule
\textbf{Collection} & \textbf{Tasks} & \textbf{\sonnet{}} & \textbf{\gptmini{}} & \textbf{\gptosslarge{}} & \textbf{\qwenmoe{}} & \textbf{\qwensmall{}} \\
\midrule
Plane partitions & 23 & 96.9 & 86.0 & 94.7 & 83.1 & 73.5 \\
Integer compositions & 62 & 87.1 & 82.4 & 80.8 & 78.8 & 41.0 \\
Integer partitions & 254 & 71.8 & 73.9 & 71.5 & 60.7 & 33.6 \\
Standard tableaux & 45 & 58.4 & 69.8 & 66.4 & 49.6 & 8.4 \\
Binary words & 96 & 84.0 & 66.4 & 83.6 & 81.8 & 34.5 \\
Permutations & 462 & 62.1 & 54.3 & 61.7 & 51.9 & 24.2 \\
\midrule
Posets & 96 & 46.5 & 47.9 & 48.8 & 36.9 & 4.8 \\
Graphs & 332 & 48.9 & 61.1 & 54.5 & 50.9 & 10.9 \\
Lattices & 40 & 38.6 & 36.4 & 29.4 & 32.4 & 5.8 \\
Dyck paths & 339 & 41.0 & 37.4 & 33.7 & 23.6 & 12.1 \\
\midrule
Set partitions & 129 & 4.1 & 3.9 & 2.9 & 3.9 & 2.7 \\
Binary trees & 52 & 3.9 & 0.0 & 0.0 & 0.0 & 0.1 \\
\bottomrule
\end{tabular}}
\caption{Strict instance accuracy (\%) by collection for five representative models spanning the closed-source production set (\sonnet{}, \gptmini{}), the open-weight frontier (\gptosslarge{}), a general-MoE open-weight system (\qwenmoe{}), and the small-model anchor (\qwensmall{}). Values in the top block ($\geq$60\% for the strongest models) span sequence- and shape-like objects; the middle block contains relational structures with high model-level variance; the bottom block is universally near zero. The full eleven-model per-collection breakdown is in App.~\ref{app:collections}.}
\label{tab:collection-strip}
\end{table}

\paragraph{Prompt length is the sharpest difficulty axis.}
Difficulty is not monotone in hidden-instance count---the 1k--9.9k bucket is \emph{easier} than the $<$100 bucket for every model---but it is sharply monotone in prompt length (\cref{tab:difficulty-axes}, top). The 24 tasks with prompts above 4{,}000 characters are a benchmark-level structural bottleneck, not a model-specific one: nearly every system collapses to single-digit macro task-instance accuracy on them, the merged \gptmini{} retry is the only exception at 15.3\% on the 4k--8k bucket (App.~\ref{app:buckets}), and all systems score 0.0\% on the single 8k+ task. Long mathematical prompts disrupt every system, regardless of provider.

\paragraph{The closed and open frontiers are complementary, not redundant.}
A coverage-controlled view of model agreement (\cref{tab:difficulty-axes}, bottom) shows that each open-weight system contributes some \emph{new} perfect tasks beyond the union of the four closed-source models. The two strongest open-weight models account for most of this complementarity (\gptosslarge{} adds 28, \gptosssmall{} adds 20); the four mid- and lower-tier OSS systems together add 29. This is the structural reason behind the +11.25\,pp ceiling lift of the eleven-model oracle reported in \cref{sec:test-time-scaling}: the gpt-oss line solves a meaningful set of tasks the four closed-source models all fail.

\begin{table}[t]
\centering
%\small
\begin{minipage}[t]{0.52\columnwidth}
\centering
\resizebox{\columnwidth}{!}{%
\begin{tabular}{lrrrr}
\toprule
\textbf{Model} & $<$2k & 2--4k & \textbf{4--8k} & 8k+ \\
\midrule
\sonnet{}     & 68.9 & 49.9 & \textbf{1.0} & 0.0 \\
\gptosslarge{} & 67.6 & 48.7 & \textbf{4.3} & 0.0 \\
\qwenmoe{}    & 59.1 & 41.8 & \textbf{9.6} & 0.0 \\
\geminiflash{} & 53.6 & 29.6 & \textbf{5.6} & 0.0 \\
\midrule
Tasks/bucket & 844 & 1{,}461 & 23 & 1 \\
\bottomrule
\end{tabular}}
\subcaption{Macro task-inst.\ accuracy (\%) by prompt-length bucket (chars).}
\label{tab:prompt-cliff}
\end{minipage}\hfill
\begin{minipage}[t]{0.44\columnwidth}
\centering
\resizebox{\columnwidth}{!}{%
\begin{tabular}{lr}
\toprule
\textbf{OSS model} & \textbf{New perfect} \\
\midrule
\gptosslarge{} & 28 \\
\gptosssmall{} & 20 \\
\qwenmoe{}     & 14 \\
\qwencoder{}   & 7 \\
\deepseekv{}   & 4 \\
\llama{}       & 3 \\
\qwensmall{}   & 1 \\
\bottomrule
\end{tabular}}
\subcaption{New perfect tasks each OSS model adds beyond the closed-source-only oracle (1{,}356 perfect tasks).}
\label{tab:marginal-oss}
\end{minipage}
\caption{Two cross-model views of difficulty. \emph{Left:} every system collapses on the 23 tasks with 4--8k-character prompts---a benchmark-level structural bottleneck (full eleven-model breakdown in App.~\ref{app:buckets}). \emph{Right:} marginal-contribution analysis---only the gpt-oss line and \qwenmoe{} contribute non-trivially to the eleven-model oracle ceiling beyond the closed-source production set.}
\label{tab:difficulty-axes}
\end{table}

\paragraph{``Reasoning ate the answer'': a budget failure mode shared across providers.}
\label{sec:budget-failure}
The gpt-oss line exposes hidden chain-of-thought via a separate \texttt{reasoning} field while reserving \texttt{content} for the final answer. At \texttt{max\_tokens=1{,}600}, 36\% of \gptosslarge{} and 46\% of \gptosssmall{} responses returned empty \texttt{content} with \texttt{finish\_reason="length"}---the entire output budget consumed by hidden reasoning before any visible code was emitted. Raising the budget to 4{,}000 leaves 24\% / 35\% of responses still hitting the cap. We score those from the reasoning trace, which typically still contains a valid \texttt{def solve} block. The same mechanism dominates \gptmini{}'s initial coverage gap: across the OpenAI Batch run we logged 889 individual batch-level failure records (the historical pre-merge total, before retries collapsed multiple failure entries onto the same task); 796 of those (89\%) carry the message ``OpenAI response did not contain any output text''---empty content with the budget exhausted on hidden reasoning. A 6{,}400-token low-effort retry recovers 567 executable programs from the 671 previously unresolved tasks, adding 62 perfect tasks and 178{,}309 correct hidden instances, while 104 tasks still emit no visible code. The recovered bucket is difficult (10.9\% covered-task and 12.5\% covered-instance accuracy), so the retry improves strict accuracy modestly but confirms that output-token budget is a first-class hyperparameter when benchmarking reasoning-trained models, regardless of provider.

\section{What helps, what hurts}
\label{sec:helps-hurts}

We isolate four common axes that affect accuracy on this benchmark. The first three are ablated directly on \qwenmoe{} (a mid-tier model where regression and improvement are both easy to measure), with self-repair additionally re-run on \gptosslarge{} as a frontier check; the fourth (cross-model oracle) extends across all eleven systems via post-hoc analysis.

\subsection{Self-repair recovers +5\,pp on a single re-prompt}
\label{sec:self-repair}

Two-turn self-repair on the full corpus---triage each turn-1 submission against its public examples, and for every task that fails any public example send a single follow-up prompt with the failing input/expected/actual triples---moves \qwenmoe{} from 37.78\,$\to$\,42.72\% task accuracy (+4.94\,pp) and 46.43\,$\to$\,51.80\% instance accuracy (+5.37\,pp) at the cost of one extra API turn on roughly half of tasks. Per-task analysis: 385 tasks improve, 274 regress (the turn-2 code introduces new errors), and 1{,}670 are unchanged; the net is +115 perfectly solved tasks and +296{,}270 correct hidden instances. The repaired \qwenmoe{} would rank between \gptmini{} and \gptosssmall{} in \cref{tab:main-results}---a single-turn execution-feedback loop is therefore a viable, cheap accuracy-recovery axis on this benchmark. Self-repair is not specific to mid-tier systems but its gain is smaller at the frontier: applying the identical protocol to \gptosslarge{} (the strongest open-weight model) raises it from 49.25\,$\to$\,50.88\% task accuracy (+1.63\,pp) and 55.13\,$\to$\,59.05\% instance accuracy (+3.92\,pp), consistent with diminishing returns. Notably the repaired \gptosslarge{} surpasses the strongest closed-source model on instance accuracy (59.05 vs.\ 56.04\%), so a single execution-feedback turn turns the open- vs.\ closed-source ``essentially tied'' state of \cref{tab:main-results} into a clean open-weight lead.

\subsection{Examples can derail classical bijections}
\label{sec:examples-derail}

We probe example-count sensitivity by regenerating prompts with $n \in \{0, 1, 3, 5\}$ public examples on a 199-task stratified subset and rerunning \qwenmoe{}. The aggregate result is non-monotone: $n{=}0$ reaches 63.1\% instance accuracy, $n{=}5$ only 49.1\%---a 14\,pp gap in the wrong direction. The effect splits sharply by family: statistics improve with examples (50.4\,$\to$\,58.2\%), maps regress sharply (70.8\,$\to$\,43.5\%). Three classical maps drive the entire map regression (\cref{tab:examples-derail}).

\begin{table}[t]
\centering
\small
\begin{tabular}{llrrr}
\toprule
\textbf{Task} & \textbf{Domain $\to$ codomain} & \textbf{$n{=}0$ acc} & \textbf{$n{=}5$ acc} & \textbf{Hidden inst.} \\
\midrule
Mp00044 & Integer partitions $\to$ Integer partitions & 100.00\% & 0.06\% & 1{,}739 \\
Mp00087 & Permutations $\to$ Permutations            & 100.00\% & 1.12\% & 61{,}219 \\
Mp00055 & Parking functions $\to$ Permutations       & 100.00\% & 12.67\% & 18{,}243 \\
\bottomrule
\end{tabular}
\caption{Three classical bijections (conjugate partition; a Foata-style cycle map; parking-function rank) where zero-example prompts reach 100\% instance accuracy but five-example prompts collapse. The $n{=}5$ code fails even on the public examples it was given (0/5 vs.\ 5/5 for the $n{=}0$ code).}
\label{tab:examples-derail}
\end{table}

We re-ran each $n{=}0$ and $n{=}5$ submission against the five public examples after canonicalisation. The $n{=}0$ codes pass 5/5 for all three tasks; the $n{=}5$ codes pass 0/5 for all three. This rules out classical few-shot overfitting (the $n{=}5$ code does not even fit the examples it was shown). The mechanism is rather a \emph{prompt-induced reasoning regression}: these maps are well-represented classical bijections (the conjugate partition map, a Foata-style cycle decomposition, and parking-function rank), and the description suffices for the model to emit a clean canonical implementation. Five worked examples---each a long combinatorial object---pull the model into an ``induce-from-examples'' mode that abandons the canonical implementation and fails. \Cref{fig:examples-derail-code} shows the actual code emitted at $n{=}0$ versus $n{=}5$ for Mp00044 (conjugate partition): the zero-shot version is the canonical 10-line implementation, the five-shot version introduces a destructive \texttt{result.pop()} and a mid-loop \texttt{break} that together produce a structurally different (and incorrect) algorithm. Examples can crowd out internalised mathematical knowledge rather than complementing it. Per-$n$ aggregates are in App.~\ref{app:passk}.

\begin{figure}[t]
\centering
\begin{minipage}[t]{0.48\columnwidth}
\centering
\textbf{$n{=}0$: 100\% on hidden, 5/5 on public}
\begin{lstlisting}[style=pyMin]
def solve(obj):
    if not obj:
        return []
    result = []
    max_len = obj[0]
    for i in range(max_len):
        count = 0
        for part in obj:
            if part > i:
                count += 1
        result.append(count)
    return result
\end{lstlisting}
\end{minipage}\hfill
\begin{minipage}[t]{0.48\columnwidth}
\centering
\textbf{$n{=}5$: 0.06\% on hidden, 0/5 on public}
\begin{lstlisting}[style=pyMin]
def solve(obj):
    if not obj:
        return []
    max_len = obj[0]
    result = [0] * max_len
    for i in range(max_len):
        while len(result) > 0 and result[-1] == 0:
            result.pop()
        for part in obj:
            if part > i:
                result[i] += 1
            else:
                break
    return result
\end{lstlisting}
\end{minipage}
\caption{\qwenmoe{} code for Mp00044 (conjugate partition; Integer partitions $\to$ Integer partitions) at zero versus five public examples. The zero-shot code emits the canonical implementation: for each $i$, the $i$-th part of the conjugate is the number of original parts strictly greater than $i$. The five-shot code introduces a destructive \texttt{result.pop()} that erases trailing zeros mid-computation and a \texttt{break} that short-circuits the inner accumulator, producing a structurally different (and incorrect) algorithm that fails on every public example it was given.}
\label{fig:examples-derail-code}
\end{figure}

\subsection{Test-time scaling: pass@5 matches the 11-model oracle ceiling}
\label{sec:test-time-scaling}

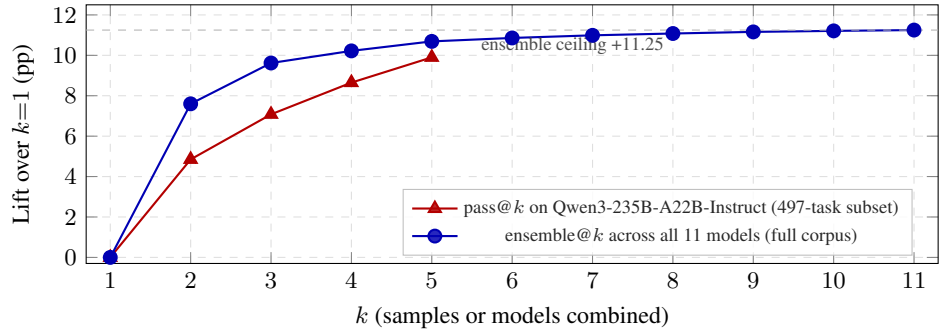
\begin{figure}[t]
\centering
\begin{tikzpicture}
\begin{axis}[
  width=0.92\columnwidth, height=5.0cm,
  xlabel={$k$ (samples or models combined)}, ylabel={Lift over $k{=}1$ (pp)},
  xmin=0.7, xmax=11.3, ymin=-0.3, ymax=12.5,
  xtick={1,2,3,4,5,6,7,8,9,10,11},
  ytick={0,2,4,6,8,10,12},
  grid=major, grid style={dashed,gray!25},
  tick label style={font=\small}, label style={font=\small},
  legend pos=south east,
  legend style={font=\scriptsize, fill=white, fill opacity=0.85, draw=gray!50},
]
\addplot[mark=triangle*, mark size=2.6pt, color=red!70!black, thick] coordinates {
  (1, 0.00) (2, 4.85) (3, 7.08) (4, 8.65) (5, 9.90)
};
\addlegendentry{pass@$k$ on \qwenmoe{} (497-task subset)}
\addplot[mark=*, mark size=2.4pt, color=blue!70!black, thick] coordinates {
  (1, 0.00) (2, 7.60) (3, 9.62) (4, 10.22) (5, 10.69)
  (6, 10.86) (7, 10.99) (8, 11.08) (9, 11.16) (10, 11.21) (11, 11.25)
};
\addlegendentry{ensemble@$k$ across all 11 models (full corpus)}
\draw[dashed,gray!60] (axis cs:0.7,11.25) -- (axis cs:11.3,11.25);
\node[anchor=west, font=\scriptsize, color=gray!60!black] at (axis cs:5.5, 10.5) {ensemble ceiling +11.25};
\end{axis}
\end{tikzpicture}
\caption{Two paths to a $\sim$10\,pp lift on \datasetname{}. Red triangles: pass@$k$ on five samples of \qwenmoe{} at temperature 0.7. Blue circles: cumulative ensemble@$k$ task accuracy as we add models in best-first order to the closed/open joint pool. Both curves saturate near $\sim$10\,pp lift over $k{=}1$---a single mid-tier model with 5-way sampling extracts headroom of the same order as an 11-model multi-provider ensemble. Adding models 5--11 to the ensemble buys only +1.03\,pp on top of the top-4 ceiling.}
\label{fig:scaling-curves}
\end{figure}

Two complementary test-time-scaling experiments expose nearly identical headroom (\cref{fig:scaling-curves}). (1) \textbf{Cross-model oracle.} A task counts as ensemble-perfect if any of the eleven evaluated models solves it perfectly. The eleven-model oracle solves 1{,}410 of 2{,}329 tasks (60.54\% task accuracy)---a +11.25\,pp ceiling above any single best model. Open- and closed-source ensembles individually reach 58.22\% (closed-only oracle, 1{,}356 tasks) and 54.87\% (OSS-only oracle, 1{,}278 tasks); \gptosslarge{} alone contributes 28 \emph{new} perfect tasks beyond the closed-source-only oracle, demonstrating that the closed and open frontiers are complementary rather than redundant. (2) \textbf{Self-consistency on a single mid-tier model.} Five independent samples of \qwenmoe{} at temperature 0.7 (top-$p$ = 0.95, max\_tokens = 1{,}600, distinct seeds) on the 497-task variance subset give per-sample task accuracies $36.98 \pm 1.51$\,pp, while pass@5 (any of the 5 perfect) reaches 46.88\%---a +9.90\,pp ceiling lift, of the same order as the 11-model oracle's +11.25\,pp. \Cref{fig:scaling-curves} plots both curves on a common axis: pass@$k$ on a single mid-tier model and ensemble@$k$ on the multi-provider mixture saturate near the same $\sim$10\,pp lift. The conservative majority bound maj@5 (perfect by $\geq$3 of 5) sits at 37.42\%, essentially identical to pass@1: the headroom comes from low-frequency lucky samples, not from a stable majority shifting position. A single mid-tier model with 5-way sampling therefore reaches the same ceiling as an 11-model multi-provider ensemble. Per-seed detail is in App.~\ref{app:passk}; full marginal-contribution and pairwise-Jaccard tables are in App.~\ref{app:oracle}.

\subsection{Greedy single-shot evaluation is stable across seeds}
\label{sec:variance}

\qwenmoe{} run at temperature 0.7 with three independent seeds on a 500-task subset has population standard deviations of 0.82\,pp on task accuracy and 1.29\,pp on instance accuracy---small relative to the 5--15\,pp inter-model gaps in \cref{tab:main-results}. The 0.92\,pp instance-accuracy gap between \gptosslarge{} and \sonnet{} is therefore inside the noise floor at this temperature, consistent with the ``essentially tied'' framing.

\section{Limitations}
\label{sec:limitations}

\begin{itemize}[leftmargin=18pt,topsep=2pt,itemsep=1pt]
  \item \textbf{Scope.} \datasetname{} targets combinatorial code synthesis from a mathematical description; results characterise this capability axis rather than general software engineering.
  \item \textbf{Single-turn protocol.} The main evaluation is one-shot; \cref{sec:self-repair} adds a single repair turn for two models. Multi-turn search, tool use, and majority voting are out of scope.
  \item \textbf{Cross-provider knobs differ.} All systems share the same prompts and sandbox, but provider-specific reasoning-effort, output-token, and response-mode controls differ; output budget is the dominant axis (\cref{sec:budget-failure}). Open-weight runs additionally use a third-party serverless API whose hardware and quantisation are not disclosed. Reported numbers are strong standardised baselines rather than tightly tuned per-provider upper bounds.
  \item \textbf{Stripped-identifier prompts.} Removing FindStat identifiers strips the most direct retrieval hooks; the underlying mathematical content remains public. A temporal slice (App.~\ref{app:temporal}) preserves the top tier and an open-book ablation (App.~\ref{app:single-model-ablations}) finds the empirical gap to source-attributed prompts is under 1\,pp.
  \item \textbf{Exact-match map evaluation.} Map outputs are scored by canonical-form exact match; we canonicalise where possible, but a small fraction of equivalent objects in non-canonical encodings can be undercounted.
  \item \textbf{Fixed evaluation corpus.} We treat the full cleaned corpus as a single test set to maximise measurement coverage; defining train/validation splits and tracking benchmark-overfitting are left to future releases.
\end{itemize}

\section{Conclusion}
\label{sec:conclusion}

\datasetname{} measures a capability current code suites under-measure: exact symbolic program induction over structured mathematical objects from a textual description and a few examples. The eleven-model evaluation supports three benchmark-level claims rather than a model-tier ranking. First, the strongest open- and closed-source models converge to within $\sim$1\,pp on this task, and the joint ceiling of all eleven models is only $\sim$10\,pp above any single model---a single mid-tier model with 5-way sampling reaches the same ceiling. Second, on classical named bijections the description already suffices: five worked examples can collapse a 100\% submission to near zero by pulling the model out of canonical-implementation mode. Third, initial coverage gaps can be reasoning-budget artefacts shared across reasoning-trained models---output-token budget is a first-class hyperparameter when benchmarking such systems. Statistic\,$\gg$\,map, near-zero set partitions and binary trees, and a sharp accuracy cliff at long prompts persist across all eleven systems and define the benchmark's structural difficulty axes. The pipeline, prompts, hidden-instance manifests, and evaluation reports are organised for release.

\bibliographystyle{plainnat}
\bibliography{references}

\clearpage
\appendix

\section{Reproducibility details}
\label{app:repro}

The supplementary materials accompanying this submission bundle a compact, anonymised reproducibility package rather than the full 21GB working directory. The archive includes: (i) the parsed task corpus (\texttt{task\_bundles.jsonl}), closed-book prompt export, and public/hidden split manifests; (ii) extracted Python submissions for all 11 systems; (iii) generation and evaluation reports for the four closed-source production systems; (iv) compact raw outputs, extracted submissions, and scoring summaries for the seven open-weight systems and ablations; (v) the benchmark package source, sandboxed evaluator, AST checker, and generation/aggregation scripts; (vi) the \LaTeX{} sources reproducing the paper. Build configuration: minimum visible pairs per task 4; public-example fraction 0.3; maximum public examples per task 5; evaluation timeout 2\,s; evaluation memory limit 256\,MB. Crawling, parsing, prompt construction, and local evaluation ran on a consumer Apple-silicon laptop in CPU-only mode; no model training was performed. Closed-model inference used provider-hosted batch services or Codex; exact accelerator type and device memory were not exposed. Open-weight inference used Together~AI's serverless API. A \texttt{README.md} in the supplementary archive documents the directory layout and reproduction commands.

\section{Per-provider model setup, knobs, and prices}
\label{app:model-setup}

The four closed-source production models use provider batch APIs: \gptmini{} via OpenAI Batch with stable low-effort reasoning (plus a 6{,}400-token retry for initial no-output tasks); \haiku{} and \sonnet{} via Anthropic Message Batches in default response mode (no explicit thinking block); \geminiflash{} via the Gemini Batch API with deterministic low-variance generation. The seven open-weight models use Together~AI serverless inference with greedy decoding (temperature 0, no system prompt, no provider-side reasoning controls), \texttt{max\_tokens=1{,}600} for non-reasoning models and \texttt{max\_tokens=4{,}000} for the gpt-oss line. Total Together~AI spend for the seven open-weight runs was \$12.96. Together list prices used: \texttt{qwen2.5-7b-turbo} \$0.18/\$0.18, \texttt{qwen3-coder-480b} \$2.00/\$2.00, \texttt{deepseek-v3.1} \$0.27/\$1.10, \texttt{qwen3-235b-tput} \$0.20/\$0.60, \texttt{llama-3.3-70b-turbo} \$0.88/\$0.88, \texttt{gpt-oss-20b} \$0.05/\$0.20, \texttt{gpt-oss-120b} \$0.15/\$0.60.

\section{Per-model main table (full columns)}
\label{app:per-model-main}

\begin{table*}[h]
\centering
\small
\resizebox{\textwidth}{!}{%
\begin{tabular}{lrrrrr}
\toprule
\textbf{Model} & \textbf{Gen.\ rate} & \textbf{Strict task} & \textbf{Strict inst.} & \textbf{Covered task} & \textbf{Covered inst.} \\
\midrule
\multicolumn{6}{l}{\textit{Closed-source production models}} \\
\sonnet{} & 99.44 & 49.29 & 56.04 & 49.57 & 56.19 \\
\gptmini{} & 95.53 & 48.35 & 52.03 & 50.61 & 53.53 \\
\haiku{} & 99.36 & 38.30 & 47.35 & 38.55 & 47.76 \\
\geminiflash{} & 99.48 & 32.33 & 39.51 & 32.50 & 39.59 \\
\midrule
\multicolumn{6}{l}{\textit{Open-weight models (Together AI serverless)}} \\
\gptosslarge{} & 99.83 & 49.25 & 55.13 & 49.33 & 55.13 \\
\gptosssmall{} & 100.00 & 43.84 & 51.27 & 43.84 & 51.27 \\
\qwenmoe{} & 100.00 & 37.78 & 46.43 & 37.78 & 46.43 \\
\qwencoder{} & 100.00 & 34.82 & 39.90 & 34.82 & 39.90 \\
\deepseekv{} & 100.00 & 31.13 & 39.16 & 31.13 & 39.16 \\
\llama{} & 100.00 & 24.52 & 32.47 & 24.52 & 32.47 \\
\qwensmall{} & 99.96 & 12.75 & 21.15 & 12.75 & 21.15 \\
\bottomrule
\end{tabular}
}
\caption{Full main-results table with generation and covered-accuracy columns. Strict task and instance accuracy use the full 2{,}329-task, 5.52M-instance denominator.}
\label{tab:appendix-main-results}
\end{table*}

\begin{table}[h]
\centering
\small
\begin{tabular}{lrrrr}
\toprule
\textbf{Model} & \textbf{Batch \$} & \textbf{Strict inst.} & \textbf{Perfect / \$} & \textbf{\$/1M correct inst.} \\
\midrule
\sonnet{} & 21.96 & 56.0 & 52.3 & 7.10 \\
\gptmini{} & 8.13 & 52.0 & 138.5 & 2.83 \\
\haiku{} & 6.12 & 47.4 & 145.7 & 2.34 \\
\geminiflash{} & 1.62 & 39.5 & 464.6 & 0.74 \\
\gptosslarge{} & 2.83 & 55.1 & 405.3 & 0.93 \\
\gptosssmall{} & 1.07 & 51.3 & 954.2 & 0.38 \\
\qwenmoe{} & 0.88 & 46.4 & 1000.0 & 0.34 \\
\qwencoder{} & 4.86 & 39.9 & 166.9 & 2.21 \\
\deepseekv{} & 1.10 & 39.2 & 659.1 & 0.51 \\
\llama{} & 1.84 & 32.5 & 310.3 & 1.03 \\
\qwensmall{} & 0.37 & 21.1 & 802.7 & 0.32 \\
\bottomrule
\end{tabular}
\caption{Cost-normalised comparison.}
\label{tab:appendix-cost}
\end{table}

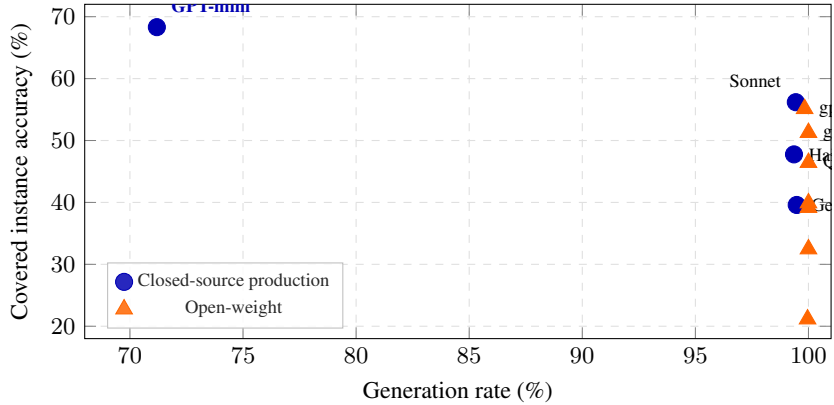
\begin{figure}[h]
\centering
\begin{tikzpicture}
\begin{axis}[
  width=0.82\textwidth, height=6.0cm,
  xlabel={Generation rate (\%)},
  ylabel={Covered instance accuracy (\%)},
  xmin=68, xmax=101, ymin=18, ymax=72,
  xtick={70,75,80,85,90,95,100},
  ytick={20,30,40,50,60,70},
  grid=major, grid style={dashed,gray!25},
  tick label style={font=\small}, label style={font=\small},
  legend pos=south west,
  legend style={font=\scriptsize, fill=white, fill opacity=0.85, draw=gray!50},
]
\addplot[only marks, mark=*, mark size=3.2pt, color=blue!70!black] coordinates {
  (99.44,56.19) (71.19,68.31) (99.36,47.76) (99.48,39.59)
};
\addlegendentry{Closed-source production}
\addplot[only marks, mark=triangle*, mark size=3.6pt, color=orange!80!red] coordinates {
  (99.83,55.13) (100.00,51.27) (100.00,46.43) (100.00,39.90)
  (100.00,39.16) (100.00,32.47) (99.96,21.15)
};
\addlegendentry{Open-weight}
\node[font=\scriptsize, above right=2pt, color=blue!70!black] at (axis cs:71.19,68.31) {\textbf{GPT-mini}};
\node[font=\scriptsize, above left=2pt] at (axis cs:99.44,56.19) {Sonnet};
\node[font=\scriptsize, right=2pt] at (axis cs:99.83,55.13) {gpt-oss-120b};
\node[font=\scriptsize, right=2pt] at (axis cs:100.00,51.27) {gpt-oss-20b};
\node[font=\scriptsize, right=2pt] at (axis cs:100.00,46.43) {Qwen3-235B};
\node[font=\scriptsize, right=2pt] at (axis cs:99.36,47.76) {Haiku};
\node[font=\scriptsize, right=2pt] at (axis cs:99.48,39.59) {Gemini};
\end{axis}
\end{tikzpicture}
\caption{Initial generation rate versus covered-instance accuracy before the 6{,}400-token \gptmini{} recovery pass. \gptmini{} is the only low-coverage point in the initial run, but conditional on producing executable code it has the highest covered-instance accuracy in the benchmark. This separates provider/output-budget coverage effects from the quality of generated programs.}
\label{fig:appendix-gen-covered}
\end{figure}

Coverage and conditional correctness are therefore not interchangeable denominators on \datasetname{}. The scatter in \cref{fig:appendix-gen-covered} shows why strict accuracy is the appropriate headline metric, while covered accuracy is diagnostically useful for identifying output-budget failures rather than mathematical weakness.

\section{Task-family breakdown}
\label{app:family}

\begin{table*}[h]
\centering
\small
\resizebox{\textwidth}{!}{%
\begin{tabular}{lrrrrrr}
\toprule
\textbf{Model} & \textbf{Stat gen.} & \textbf{Stat strict inst.} & \textbf{Stat covered inst.} & \textbf{Map gen.} & \textbf{Map strict inst.} & \textbf{Map covered inst.} \\
\midrule
\sonnet{} & 99.15 & 65.82 & 66.24 & 95.24 & 50.79 & 50.82 \\
\gptmini{} & 95.38 & 66.15 & 67.98 & 96.43 & 44.44 & 45.75 \\
\haiku{} & 99.35 & 57.22 & 57.64 & 99.40 & 42.04 & 42.44 \\
\geminiflash{} & 99.45 & 46.97 & 47.20 & 99.70 & 35.50 & 35.52 \\
\gptosslarge{} & 99.80 & 64.02 & 64.04 & 100.00 & 50.34 & 50.34 \\
\gptosssmall{} & 100.00 & 57.89 & 57.89 & 100.00 & 47.71 & 47.71 \\
\qwenmoe{} & 100.00 & 57.18 & 57.18 & 100.00 & 40.65 & 40.65 \\
\qwencoder{} & 100.00 & 53.04 & 53.04 & 100.00 & 32.84 & 32.84 \\
\deepseekv{} & 100.00 & 48.74 & 48.74 & 100.00 & 34.00 & 34.00 \\
\llama{} & 100.00 & 42.97 & 42.97 & 100.00 & 26.83 & 26.83 \\
\qwensmall{} & 100.00 & 27.02 & 27.02 & 99.70 & 17.99 & 18.00 \\
\bottomrule
\end{tabular}
}
\caption{Per-family statistic and map breakdown. The map ratio for \qwencoder{} (32.84 / 53.04 = 0.62) is the lowest in the paper.}
\label{tab:appendix-family}
\end{table*}

\section{Failure-mode breakdown}
\label{app:failures}

\begin{table}[h]
\centering
\small
\begin{tabular}{lrrrrr}
\toprule
\textbf{Model} & \textbf{Gen.\ miss} & \textbf{Perfect} & \textbf{Partial} & \textbf{Zero-pass} & \textbf{Syntax err.} \\
\midrule
\gptmini{} & 104 & 1{,}126 & 633 & 466 & 77 \\
\haiku{} & 15 & 892 & 907 & 515 & 53 \\
\sonnet{} & 13 & 1{,}148 & 511 & 657 & 115 \\
\geminiflash{} & 12 & 753 & 569 & 995 & 341 \\
\gptosslarge{} & 4 & 1{,}147 & 295 & 883 & 548 \\
\gptosssmall{} & 0 & 1{,}021 & 241 & 1{,}067 & 805 \\
\qwenmoe{} & 0 & 880 & 849 & 600 & 77 \\
\qwencoder{} & 0 & 811 & 983 & 535 & 78 \\
\deepseekv{} & 0 & 725 & 1{,}014 & 590 & 30 \\
\llama{} & 0 & 571 & 1{,}179 & 579 & 3 \\
\qwensmall{} & 1 & 297 & 1{,}107 & 924 & 20 \\
\bottomrule
\end{tabular}
\caption{Task-level failure decomposition. ``Perfect'', ``Partial'', and ``Zero-pass'' are computed over generated tasks only. ``Gen.\ miss'' is benchmark tasks with no executable submission. The inflated syntax-error counts for the gpt-oss line include reasoning-only responses (\cref{sec:budget-failure}). For \gptmini{}, the initial run left 671 unique tasks unresolved; a 6{,}400-token retry recovered 567 executable programs, leaving 104 no-output tasks in the final merged run.}
\label{tab:appendix-failures}
\end{table}

\paragraph{Anti-cheating audit.}
The AST checker scans for large literal tables, branches on exact public-example strings, repeated string literals, and disallowed imports. In the four main runs we found no generated task flagged for large literals or explicit example-branching; the only non-syntax suspicious tasks were one GPT and one Sonnet submission that imported \texttt{sys}, both rejected by the sandbox. Most remaining ``suspicious'' flags are syntax-invalid generations, not lookup-table behaviour.

\section{Per-collection $\times$ per-model breakdown}
\label{app:collections}

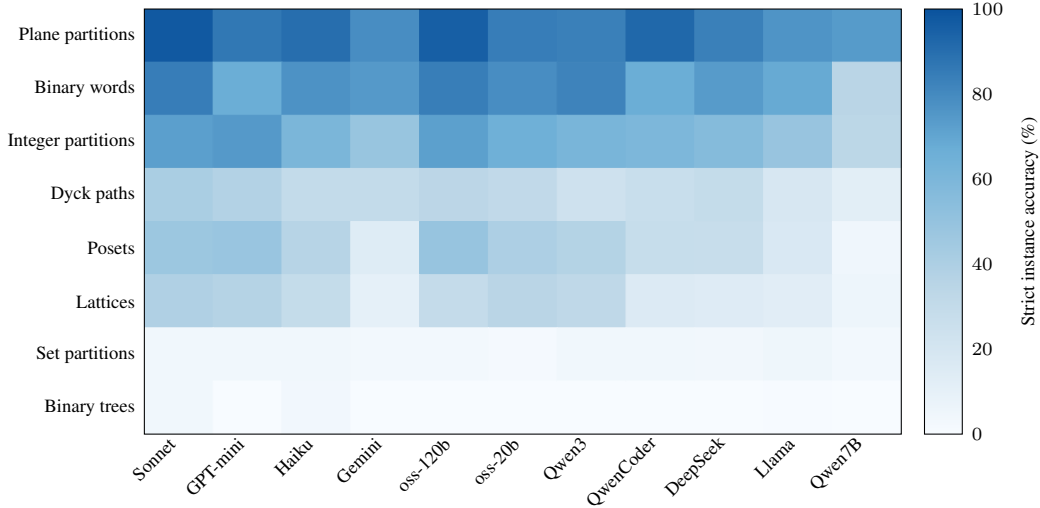
\begin{figure*}[h]
\centering
\begin{tikzpicture}
\begin{axis}[
  width=0.83\textwidth, height=7.2cm,
  enlargelimits=false,
  axis on top,
  colorbar,
  colormap={findstatheat}{rgb255(0cm)=(247,251,255); rgb255(1cm)=(189,215,231); rgb255(2cm)=(107,174,214); rgb255(3cm)=(8,81,156)},
  point meta min=0,
  point meta max=100,
  colorbar style={
    ylabel={Strict instance accuracy (\%)},
    ytick={0,20,40,60,80,100},
    tick label style={font=\scriptsize},
    label style={font=\scriptsize}
  },
  xmin=-0.5, xmax=10.5,
  ymin=-0.5, ymax=7.5,
  xtick={0,1,2,3,4,5,6,7,8,9,10},
  xticklabels={Sonnet,GPT-mini,Haiku,Gemini,oss-120b,oss-20b,Qwen3,QwenCoder,DeepSeek,Llama,Qwen7B},
  ytick={0,1,2,3,4,5,6,7},
  yticklabels={Binary trees,Set partitions,Lattices,Posets,Dyck paths,Integer partitions,Binary words,Plane partitions},
  x tick label style={rotate=45, anchor=east, font=\scriptsize},
  y tick label style={font=\scriptsize},
  tick style={draw=none},
]
\addplot[
  matrix plot*,
  mesh/cols=11,
  point meta=explicit,
] table [x=x, y=y, meta=acc] {
x y acc
0 7 96.9
1 7 86.0
2 7 89.3
3 7 78.2
4 7 94.7
5 7 84.0
6 7 83.1
7 7 91.6
8 7 83.1
9 7 75.9
10 7 73.5
0 6 84.0
1 6 66.4
2 6 76.5
3 6 74.0
4 6 83.6
5 6 78.2
6 6 81.8
7 6 66.7
8 6 73.3
9 6 68.0
10 6 34.5
0 5 71.8
1 5 73.9
2 5 60.2
3 5 48.1
4 5 71.5
5 5 64.9
6 5 60.7
7 5 59.2
8 5 56.1
9 5 48.4
10 5 33.6
0 4 40.9
1 4 37.4
2 4 29.7
3 4 29.7
4 4 33.7
5 4 31.1
6 4 23.6
7 4 27.0
8 4 29.1
9 4 18.4
10 4 12.1
0 3 46.5
1 3 47.9
2 3 35.9
3 3 14.2
4 3 48.8
5 3 39.6
6 3 36.9
7 3 27.9
8 3 27.8
9 3 17.3
10 3 4.8
0 2 38.6
1 2 36.4
2 2 29.0
3 2 10.1
4 2 29.4
5 2 34.6
6 2 32.4
7 2 15.4
8 2 14.4
9 2 12.7
10 2 5.8
0 1 4.1
1 1 3.9
2 1 4.0
3 1 2.8
4 1 2.9
5 1 1.6
6 1 3.9
7 1 3.9
8 1 3.7
9 1 5.0
10 1 2.7
0 0 3.9
1 0 0.0
2 0 3.4
3 0 0.0
4 0 0.0
5 0 0.0
6 0 0.0
7 0 0.0
8 0 0.0
9 0 0.6
10 0 0.1
};
\end{axis}
\end{tikzpicture}
\caption{Model $\times$ collection heatmap for the eight representative collections reported for all eleven systems. Color encodes strict instance accuracy (\%): plane partitions, binary words, and integer partitions are broadly easier, while set partitions and binary trees remain near zero.}
\label{fig:appendix-collection-heatmap}
\end{figure*}

The heatmap in \cref{fig:appendix-collection-heatmap} expands the representative collection strip in \cref{tab:collection-strip} to all eleven evaluated systems on the shared collection rows; exact values are given in \cref{tab:appendix-collections-closed,tab:appendix-collections-oss}. It makes the main qualitative pattern visually explicit: the benchmark contains collections where many systems are competent, collections where model choice matters sharply, and collections where every current system is brittle.

\begin{table}[h]
\centering
\scriptsize
\resizebox{\columnwidth}{!}{%
\begin{tabular}{L{2.6cm}rrrrr}
\toprule
\textbf{Collection} & \textbf{Tasks} & \textbf{\gptmini{}} & \textbf{\haiku{}} & \textbf{\sonnet{}} & \textbf{\geminiflash{}} \\
\midrule
Graphs & 332 & 61.1 & 51.0 & 48.9 & 16.6 \\
Posets & 96 & 47.9 & 35.9 & 46.5 & 14.2 \\
Lattices & 40 & 36.4 & 29.0 & 38.6 & 10.1 \\
Integer partitions & 254 & 73.9 & 60.2 & 71.8 & 48.1 \\
Binary words & 96 & 66.4 & 76.5 & 84.0 & 74.0 \\
Dyck paths & 339 & 37.4 & 29.7 & 40.9 & 29.7 \\
Plane partitions & 23 & 86.0 & 89.3 & 96.9 & 78.2 \\
Set partitions & 129 & 3.9 & 4.0 & 4.1 & 2.8 \\
Binary trees & 52 & 0.0 & 3.4 & 3.9 & 0.0 \\
\bottomrule
\end{tabular}
}
\caption{Closed-source instance accuracy (\%) by representative collection.}
\label{tab:appendix-collections-closed}
\end{table}

\begin{table*}[h]
\centering
\scriptsize
\resizebox{\textwidth}{!}{%
\begin{tabular}{L{2.6cm}rrrrrrrr}
\toprule
\textbf{Collection} & \textbf{Tasks} & \textbf{\gptosslarge{}} & \textbf{\gptosssmall{}} & \textbf{\qwenmoe{}} & \textbf{\qwencoder{}} & \textbf{\deepseekv{}} & \textbf{\llama{}} & \textbf{\qwensmall{}} \\
\midrule
Plane partitions & 23 & 94.7 & 84.0 & 83.1 & 91.6 & 83.1 & 75.9 & 73.5 \\
Integer compositions & 62 & 80.8 & 65.3 & 78.8 & 66.7 & 64.3 & 52.4 & 41.0 \\
Standard tableaux & 45 & 66.4 & 63.2 & 49.6 & 45.3 & 37.6 & 30.8 & 8.4 \\
Integer partitions & 254 & 71.5 & 64.9 & 60.7 & 59.2 & 56.1 & 48.4 & 33.6 \\
Binary words & 96 & 83.6 & 78.2 & 81.8 & 66.7 & 73.3 & 68.0 & 34.5 \\
Binary trees & 52 & 0.0 & 0.0 & 0.0 & 0.0 & 0.0 & 0.6 & 0.1 \\
Set partitions & 129 & 2.9 & 1.6 & 3.9 & 3.9 & 3.7 & 5.0 & 2.7 \\
Dyck paths & 339 & 33.7 & 31.1 & 23.6 & 27.0 & 29.1 & 18.4 & 12.1 \\
Lattices & 40 & 29.4 & 34.6 & 32.4 & 15.4 & 14.4 & 12.7 & 5.8 \\
Posets & 96 & 48.8 & 39.6 & 36.9 & 27.9 & 27.8 & 17.3 & 4.8 \\
\bottomrule
\end{tabular}
}
\caption{Open-weight instance accuracy (\%) by representative collection.}
\label{tab:appendix-collections-oss}
\end{table*}

\section{Map-codomain extremes}
\label{app:codomain}

\begin{table*}[h]
\centering
\scriptsize
\resizebox{\textwidth}{!}{%
\begin{tabular}{L{3.0cm}r|rr|rr|rr|rr}
\toprule
\multirow{2}{*}{\textbf{Codomain}} & \multirow{2}{*}{\textbf{Tasks}} & \multicolumn{2}{c|}{\textbf{\gptmini{}}} & \multicolumn{2}{c|}{\textbf{\haiku{}}} & \multicolumn{2}{c|}{\textbf{\sonnet{}}} & \multicolumn{2}{c}{\textbf{\geminiflash{}}} \\
 &  & \textbf{Gen.} & \textbf{Strict inst.} & \textbf{Gen.} & \textbf{Strict inst.} & \textbf{Gen.} & \textbf{Strict inst.} & \textbf{Gen.} & \textbf{Strict inst.} \\
\midrule
Standard tableaux & 12 & 100.0 & 87.2 & 100.0 & 80.5 & 100.0 & 79.0 & 100.0 & 83.1 \\
Binary words & 27 & 88.9 & 76.7 & 100.0 & 76.4 & 100.0 & 94.8 & 100.0 & 77.1 \\
Integer compositions & 25 & 96.0 & 73.9 & 100.0 & 66.2 & 100.0 & 72.8 & 100.0 & 57.3 \\
Parking functions & 10 & 100.0 & 69.5 & 100.0 & 61.3 & 100.0 & 69.4 & 100.0 & 49.2 \\
Signed permutations & 13 & 92.3 & 66.3 & 100.0 & 58.8 & 100.0 & 66.9 & 100.0 & 52.8 \\
\midrule
Set partitions & 21 & 95.2 & 0.9 & 100.0 & 0.0 & 100.0 & 0.8 & 100.0 & 0.0 \\
Binary trees & 12 & 91.7 & 0.0 & 100.0 & 0.0 & 91.7 & 37.2 & 100.0 & 0.0 \\
Lattices & 10 & 100.0 & 0.0 & 100.0 & 0.0 & 100.0 & 0.0 & 100.0 & 0.0 \\
Posets & 14 & 100.0 & 2.9 & 100.0 & 3.6 & 100.0 & 0.0 & 100.0 & 3.5 \\
Graphs & 20 & 100.0 & 11.8 & 100.0 & 11.6 & 100.0 & 2.2 & 95.0 & 2.7 \\
\bottomrule
\end{tabular}
}
\caption{Representative map-codomain breakdown for codomains with at least 10 map tasks.}
\label{tab:appendix-codomains}
\end{table*}

\section{Prompt-length and instance-bucket sensitivity}
\label{app:buckets}

\begin{table*}[h]
\centering
\scriptsize
\resizebox{\textwidth}{!}{%
\begin{tabular}{L{2.2cm}|rrrr|rrrr}
\toprule
\multirow{2}{*}{\textbf{Model}} & \multicolumn{4}{c|}{\textbf{Hidden-instance bucket (macro task-inst.\ \%)}} & \multicolumn{4}{c}{\textbf{Prompt-length bucket (macro task-inst.\ \%)}} \\
 & \textbf{<100} & \textbf{100--999} & \textbf{1k--9.9k} & \textbf{10k+} & \textbf{<2k} & \textbf{2k--4k} & \textbf{4k--8k} & \textbf{8k+} \\
\midrule
\sonnet{} & 43.2 & 51.2 & 60.2 & 55.7 & 68.9 & 49.9 & 1.0 & 0.0 \\
\gptmini{} & 54.5 & 56.8 & 59.9 & 51.6 & 66.5 & 54.4 & 15.3 & 0.0 \\
\haiku{} & 37.1 & 45.7 & 51.9 & 42.9 & 58.8 & 43.6 & 10.1 & 0.0 \\
\geminiflash{} & 25.8 & 32.7 & 42.1 & 36.8 & 53.6 & 29.6 & 5.6 & 0.0 \\
\gptosslarge{} & 41.0 & 49.7 & 59.3 & 54.2 & 67.6 & 48.7 & 4.3 & 0.0 \\
\gptosssmall{} & 30.5 & 43.2 & 53.4 & 52.3 & 60.6 & 42.7 & 9.3 & 0.0 \\
\qwenmoe{} & 34.1 & 42.6 & 52.0 & 43.0 & 59.1 & 41.8 & 9.6 & 0.0 \\
\qwencoder{} & 32.4 & 43.1 & 47.8 & 33.6 & 55.6 & 39.3 & 10.5 & 0.0 \\
\deepseekv{} & 29.3 & 38.6 & 44.5 & 34.4 & 51.1 & 36.2 & 11.1 & 0.0 \\
\llama{} & 22.4 & 32.7 & 38.2 & 27.0 & 46.7 & 28.9 & 4.5 & 0.0 \\
\qwensmall{} & 13.3 & 23.4 & 23.1 & 16.9 & 33.3 & 16.2 & 1.0 & 0.0 \\
\bottomrule
\end{tabular}
}
\caption{Macro per-task hidden-instance accuracy by task size and prompt length. The first size bucket contains tasks with fewer than 100 hidden instances. Prompt-length variation is highly imbalanced: 844 tasks have prompts $<$2k characters, 1{,}461 in 2k--4k, 23 in 4k--8k, 1 above 8k. The \gptmini{} row uses the merged low-effort run with the 6{,}400-token recovery pass.}
\label{tab:appendix-buckets}
\end{table*}

\section{Recent-task temporal slice}
\label{app:temporal}

\begin{table}[h]
\centering
\small
\begin{tabular}{lrrrr}
\toprule
\textbf{Model} & \textbf{Gen.} & \textbf{Strict task} & \textbf{Strict inst.} & \textbf{Macro} \\
\midrule
\sonnet{} & 99.7 & 36.2 & 53.0 & 44.3 \\
\gptmini{} & 89.6 & 38.3 & 53.3 & 52.3 \\
\haiku{} & 99.3 & 27.2 & 42.6 & 40.5 \\
\geminiflash{} & 98.7 & 22.5 & 35.6 & 29.5 \\
\bottomrule
\end{tabular}
\caption{Temporal slice on tasks with FindStat page updates in 2024--2026 (298 tasks). The closed-source top tier remains similar to the full corpus; after the 6{,}400-token recovery pass, \gptmini{} slightly leads this small slice on strict instance accuracy.}
\label{tab:appendix-temporal}
\end{table}

\section{Cross-model oracle (full)}
\label{app:oracle}

\paragraph{Marginal contribution to the closed-source oracle.}
Each OSS model adds $k$ \emph{new} perfect tasks beyond the closed-source-only oracle (1{,}356 perfect tasks): \gptosslarge{} 28, \gptosssmall{} 20, \qwenmoe{} 14, \qwencoder{} 7, \deepseekv{} 4, \llama{} 3, \qwensmall{} 1. The gpt-oss line is therefore complementary to the closed-source production set, not redundant with it.

\paragraph{Pairwise Jaccard overlap.}
On perfect-task sets (top pairs): \sonnet{}\,$\cap$\,\gptosslarge{} = 1{,}004 / union 1{,}291 (Jaccard 0.78), with 144 tasks unique to Sonnet and 143 unique to gpt-oss-120b. The two strongest models in the paper still disagree on $\sim$22\% of their perfect-task union.

\paragraph{$k$-of-7 OSS agreement.}
Of 1{,}278 tasks perfect by at least one OSS model: 156 by exactly one, 183 by two, 156 by three, 146 by four, 180 by five, 221 by six, 236 by all seven; tasks perfect by $\geq$4 OSS models number 783 (33.6\% of the corpus).

\section{Pass@k and example-count details}
\label{app:passk}

\paragraph{Per-seed pass@1 and pass@k.}
Five \qwenmoe{} samples at $T{=}0.7$ (top-$p$ = 0.95, max\_tokens = 1{,}600, seeds 11/22/33/44/55) on the 497-task variance subset:

\begin{table}[h]
\centering
\small
\begin{tabular}{lrr}
\toprule
\textbf{Aggregate} & \textbf{Strict task (\%)} & \textbf{Strict inst.\ (\%)} \\
\midrule
sample 1 (seed 11) & 34.41 & 50.53 \\
sample 2 (seed 22) & 36.62 & 50.03 \\
sample 3 (seed 33) & 37.42 & 50.49 \\
sample 4 (seed 44) & 37.42 & 54.87 \\
sample 5 (seed 55) & 39.03 & 51.86 \\
\midrule
pass@1 (mean $\pm$ std) & 36.98 $\pm$ 1.51 & 51.55 $\pm$ 1.77 \\
\textbf{pass@5} (any of 5) & \textbf{46.88} & --- \\
\textbf{maj@5} ($\geq$3 of 5) & \textbf{37.42} & --- \\
\bottomrule
\end{tabular}
\caption{Per-seed and aggregate pass@k for \qwenmoe{}. Of 233 tasks perfect by at least one of 5 samples: 129 are perfect by all 5, 40 by 4 of 5, 17 by 3 of 5, 16 by 2 of 5, and 31 by exactly one.}
\label{tab:appendix-passk}
\end{table}

\paragraph{Per-$n$ example-count aggregates.}
Five-way example-count ablation on \qwenmoe{} on the 199-task stratified subset:

\begin{table}[h]
\centering
\small
\begin{tabular}{rrrrr}
\toprule
\textbf{$n$} & \textbf{Strict task (\%)} & \textbf{Strict inst.\ (\%)} & \textbf{Stat inst.\ (\%)} & \textbf{Map inst.\ (\%)} \\
\midrule
0 & 35.18 & 63.12 & 50.38 & 70.84 \\
1 & 40.20 & 53.69 & 58.63 & 50.69 \\
3 & 39.20 & 49.95 & 58.68 & 44.65 \\
5 & 38.69 & 49.05 & 58.20 & 43.50 \\
\bottomrule
\end{tabular}
\caption{Public-example-count ablation. Statistics improve with examples; maps regress sharply (\cref{sec:examples-derail}).}
\label{tab:appendix-examplecount}
\end{table}

\section{Self-repair on the open-weight frontier}
\label{app:single-model-ablations}

To check whether the +5\,pp self-repair gain reported in \cref{sec:self-repair} for \qwenmoe{} extends to the strongest open-weight model, we re-ran the identical two-turn protocol with \gptosslarge{} as both turn-1 and turn-2 model (\texttt{max\_tokens=4{,}000} on the turn-2 call) and scored the merged turn-1$+$turn-2 submission against the same 5.52M-instance hidden corpus. Numbers: 49.25\,$\to$\,50.88\% task accuracy (+1.63\,pp) and 55.13\,$\to$\,59.05\% instance accuracy (+3.92\,pp); 1{,}147 perfect tasks $\to$ 1{,}185 (+38); 3{,}041{,}273 $\to$ 3{,}257{,}571 correct hidden instances (+216{,}298). The gain is positive but smaller than the +5.37\,pp instance-accuracy gain on \qwenmoe{}, consistent with diminishing returns at the frontier; the repaired \gptosslarge{} also strictly leads the strongest closed-source model on instance accuracy (59.05 vs.\ \sonnet{}'s 56.04\%). Two further single-model ablations confirm related trends. \emph{Reasoning-budget scaling on \gptosslarge{}:} re-running the 557 length-capped tasks at \texttt{max\_tokens=8{,}000} and merging (8k output where it parses, original turn-1 code otherwise) raises strict accuracy from 49.25\,/\,55.13 to 52.08\,/\,58.59\% (+2.83\,/\,+3.46\,pp), 1{,}147\,$\to$\,1{,}213 perfect tasks (+66); like the self-repair pass, the rescued submission strictly surpasses the strongest closed-source model on instance accuracy. \emph{Open-book vs.\ closed-book on \qwenmoe{}:} on the 199-task example-ablation subset, the closed-book $n{=}5$ baseline reaches 38.69\,/\,49.05\% task / instance accuracy and the open-book $n{=}5$ variant (FindStat identifiers, URLs, and source branding retained) reaches 39.70\,/\,49.85\%---a +1.01\,pp task and +0.80\,pp instance-accuracy gap, confirming that the FindStat-identifier stripping accounts for less than 1\,pp on instance accuracy and that the closed-book/open-book distinction is methodological rather than a major contamination control on this corpus.

\section{GPT-mini medium-reasoning ablation}
\label{app:gpt-medium}

A within-model ablation for \gptmini{} using medium-effort reasoning was collected in three stages---a 539-task Codex-plan run, a 718-task OpenAI Batch run with \texttt{max\_output\_tokens=1{,}600}, and a 150-task OpenAI Batch recovery run with \texttt{max\_output\_tokens=3{,}200}. The merged medium run reaches 60.41\% generation rate, 42.94\% strict task accuracy, 49.81\% strict instance accuracy, 71.07\% covered task accuracy, and 81.29\% covered instance accuracy. Relative to the low-effort main run after the 6{,}400-token recovery pass, medium reasoning has lower strict task and instance accuracy (42.94 / 49.81 vs.\ 48.35 / 52.03) but substantially higher covered accuracy on the smaller subset it answers. We treat this as a supplementary ablation rather than a main-table leaderboard row because the serving backends and response budgets differ.

\section{Asset provenance, licensing, and release status}
\label{app:licensing}

\paragraph{Source data: FindStat.}
The benchmark is derived from FindStat~\citep{findstat}, a public-facing combinatorial database whose page content (collection definitions, statistic and map descriptions, visible object--integer and object--object image pairs) is openly accessible at \url{https://www.findstat.org/}. We crawl static pages with rate-limited, exponentially-backed-off polite requests and cache HTML on disk; no authenticated or paywalled content is retrieved. Where applicable we cite FindStat by name in any derived artefact, and the released supplementary materials retain attribution metadata for every parsed page.

\paragraph{Derived benchmark.}
The parsed JSONL artefacts, closed-book prompts, public-hidden splits, and evaluation reports are derivative works of the FindStat content. The supplementary release provides benchmark data and reports under CC BY 4.0 for research and review use, with FindStat attribution retained in the task metadata and subject to upstream FindStat terms for source content. The release archive includes a provenance note for every collection, statistic, and map id present in the corpus.

\paragraph{Model outputs and submissions.}
The eleven extracted Python submissions, compact raw output exports, and per-model evaluation summaries are released as research artefacts alongside the corpus. We retain the original provider and model identifier in every record. Use of these outputs is subject to the upstream provider terms of service in effect at run time (OpenAI Batch API, Anthropic Message Batches, Google Gemini Batch API, Together~AI serverless inference); the provider terms govern redistribution of model-generated content beyond the release of these specific outputs.

\paragraph{Evaluator and pipeline code.}
The crawler, parser, prompt builder, sandboxed evaluator, and AST checker are included in the supplementary archive under the MIT License. Reviewers can reproduce the reported scores by running the supplied scripts against the released artefacts.

\paragraph{Release restrictions and risks.}
Because the released objects are public combinatorial structures rather than personal data, scraped media, or user-generated content, we do not impose access controls. The main release risks are (i) terms-of-use compliance with the upstream FindStat site, addressed through attribution and research-use framing, and (ii) overfitting of future models to the fixed hidden-test split, which we mitigate by reporting denominator-sensitive metrics and recent-task slices so that future work can construct disjoint evaluation subsets.

\section{Prompt template summary}
\label{app:prompt-template}

The closed-book prompt template is:

\begin{enumerate}[leftmargin=18pt,topsep=2pt,itemsep=1pt]
  \item Instruct the model to output only Python code defining \texttt{solve(obj)}.
  \item State whether the output is an integer or a canonicalised object.
  \item Provide collection / domain / codomain names and definitions.
  \item Provide encoding notes and optional enumerative or map-property notes.
  \item Provide the task description and up to five public examples.
\end{enumerate}

The closed-book template differs from the open-book variant only by removing direct source attribution (FindStat IDs, URLs, source branding) from the prompt text.

\clearpage
\section*{NeurIPS Paper Checklist}

\begin{enumerate}

\item {\bf Claims}
    \item[] Question: Do the main claims made in the abstract and introduction accurately reflect the paper's contributions and scope?
    \item[] Answer: \answerYes{}
    \item[] Justification: The abstract and introduction describe the benchmark construction, the single-turn evaluation protocol, the eleven-model comparison, the four cross-model ablations / analyses (self-repair, example-count, test-time scaling, variance), and the scoped claims about combinatorial code synthesis rather than general software engineering ability (Abstract; Sections~1, 4, and 6).
    \item[] Guidelines:
    \begin{itemize}
        \item The answer \answerNA{} means that the abstract and introduction do not include the claims made in the paper.
        \item The abstract and/or introduction should clearly state the claims made, including the contributions made in the paper and important assumptions and limitations. A \answerNo{} or \answerNA{} answer to this question will not be perceived well by the reviewers. 
        \item The claims made should match theoretical and experimental results, and reflect how much the results can be expected to generalize to other settings. 
        \item It is fine to include aspirational goals as motivation as long as it is clear that these goals are not attained by the paper. 
    \end{itemize}

\item {\bf Limitations}
    \item[] Question: Does the paper discuss the limitations of the work performed by the authors?
    \item[] Answer: \answerYes{}
    \item[] Justification: The paper includes a dedicated ``Limitations'' section discussing scope, provider-setting mismatch, single-turn evaluation, possible contamination, fixed test-only release, open-weight inference variability, exact-match harshness, and benchmark-overfitting risk (Section 7).
    \item[] Guidelines:
    \begin{itemize}
        \item The answer \answerNA{} means that the paper has no limitation while the answer \answerNo{} means that the paper has limitations, but those are not discussed in the paper. 
        \item The authors are encouraged to create a separate ``Limitations'' section in their paper.
        \item The paper should point out any strong assumptions and how robust the results are to violations of these assumptions (e.g., independence assumptions, noiseless settings, model well-specification, asymptotic approximations only holding locally). The authors should reflect on how these assumptions might be violated in practice and what the implications would be.
        \item The authors should reflect on the scope of the claims made, e.g., if the approach was only tested on a few datasets or with a few runs. In general, empirical results often depend on implicit assumptions, which should be articulated.
        \item The authors should reflect on the factors that influence the performance of the approach. For example, a facial recognition algorithm may perform poorly when image resolution is low or images are taken in low lighting. Or a speech-to-text system might not be used reliably to provide closed captions for online lectures because it fails to handle technical jargon.
        \item The authors should discuss the computational efficiency of the proposed algorithms and how they scale with dataset size.
        \item If applicable, the authors should discuss possible limitations of their approach to address problems of privacy and fairness.
        \item While the authors might fear that complete honesty about limitations might be used by reviewers as grounds for rejection, a worse outcome might be that reviewers discover limitations that aren't acknowledged in the paper. The authors should use their best judgment and recognize that individual actions in favor of transparency play an important role in developing norms that preserve the integrity of the community. Reviewers will be specifically instructed to not penalize honesty concerning limitations.
    \end{itemize}

\item {\bf Theory assumptions and proofs}
    \item[] Question: For each theoretical result, does the paper provide the full set of assumptions and a complete (and correct) proof?
    \item[] Answer: \answerNA{}
    \item[] Justification: The paper is an empirical benchmark and evaluation study; it does not present theorems or formal proofs.
    \item[] Guidelines:
    \begin{itemize}
        \item The answer \answerNA{} means that the paper does not include theoretical results. 
        \item All the theorems, formulas, and proofs in the paper should be numbered and cross-referenced.
        \item All assumptions should be clearly stated or referenced in the statement of any theorems.
        \item The proofs can either appear in the main paper or the supplemental material, but if they appear in the supplemental material, the authors are encouraged to provide a short proof sketch to provide intuition. 
        \item Inversely, any informal proof provided in the core of the paper should be complemented by formal proofs provided in appendix or supplemental material.
        \item Theorems and Lemmas that the proof relies upon should be properly referenced. 
    \end{itemize}

    \item {\bf Experimental result reproducibility}
    \item[] Question: Does the paper fully disclose all the information needed to reproduce the main experimental results of the paper to the extent that it affects the main claims and/or conclusions of the paper (regardless of whether the code and data are provided or not)?
    \item[] Answer: \answerYes{}
    \item[] Justification: The benchmark construction, prompt format, evaluator, metrics, model setup, and ablation protocol are specified in the main paper and appendix, including exact task counts, hidden-instance counts, and run settings that affect the reported conclusions (Sections 2--4; Appendix A).
    \item[] Guidelines:
    \begin{itemize}
        \item The answer \answerNA{} means that the paper does not include experiments.
        \item If the paper includes experiments, a \answerNo{} answer to this question will not be perceived well by the reviewers: Making the paper reproducible is important, regardless of whether the code and data are provided or not.
        \item If the contribution is a dataset and\slash or model, the authors should describe the steps taken to make their results reproducible or verifiable. 
        \item Depending on the contribution, reproducibility can be accomplished in various ways. For example, if the contribution is a novel architecture, describing the architecture fully might suffice, or if the contribution is a specific model and empirical evaluation, it may be necessary to either make it possible for others to replicate the model with the same dataset, or provide access to the model. In general. releasing code and data is often one good way to accomplish this, but reproducibility can also be provided via detailed instructions for how to replicate the results, access to a hosted model (e.g., in the case of a large language model), releasing of a model checkpoint, or other means that are appropriate to the research performed.
        \item While NeurIPS does not require releasing code, the conference does require all submissions to provide some reasonable avenue for reproducibility, which may depend on the nature of the contribution. For example
        \begin{enumerate}
            \item If the contribution is primarily a new algorithm, the paper should make it clear how to reproduce that algorithm.
            \item If the contribution is primarily a new model architecture, the paper should describe the architecture clearly and fully.
            \item If the contribution is a new model (e.g., a large language model), then there should either be a way to access this model for reproducing the results or a way to reproduce the model (e.g., with an open-source dataset or instructions for how to construct the dataset).
            \item We recognize that reproducibility may be tricky in some cases, in which case authors are welcome to describe the particular way they provide for reproducibility. In the case of closed-source models, it may be that access to the model is limited in some way (e.g., to registered users), but it should be possible for other researchers to have some path to reproducing or verifying the results.
        \end{enumerate}
    \end{itemize}

\item {\bf Open access to data and code}
    \item[] Question: Does the paper provide open access to the data and code, with sufficient instructions to faithfully reproduce the main experimental results, as described in supplemental material?
    \item[] Answer: \answerYes{}
    \item[] Justification: The submission ships a compact supplementary archive containing the benchmark pipeline (crawler, parser, prompt builder, sandboxed evaluator, AST-checker), the parsed task corpus and prompt exports, public/hidden split manifests, extracted submissions for all eleven systems, compact raw-output and scoring summaries where size permits, aggregation/ablation scripts, and a \texttt{README.md} documenting the reproduction commands (Appendix~A).
    \item[] Guidelines:
    \begin{itemize}
        \item The answer \answerNA{} means that paper does not include experiments requiring code.
        \item Please see the NeurIPS code and data submission guidelines (\url{https://neurips.cc/public/guides/CodeSubmissionPolicy}) for more details.
        \item While we encourage the release of code and data, we understand that this might not be possible, so \answerNo{} is an acceptable answer. Papers cannot be rejected simply for not including code, unless this is central to the contribution (e.g., for a new open-source benchmark).
        \item The instructions should contain the exact command and environment needed to run to reproduce the results. See the NeurIPS code and data submission guidelines (\url{https://neurips.cc/public/guides/CodeSubmissionPolicy}) for more details.
        \item The authors should provide instructions on data access and preparation, including how to access the raw data, preprocessed data, intermediate data, and generated data, etc.
        \item The authors should provide scripts to reproduce all experimental results for the new proposed method and baselines. If only a subset of experiments are reproducible, they should state which ones are omitted from the script and why.
        \item At submission time, to preserve anonymity, the authors should release anonymized versions (if applicable).
        \item Providing as much information as possible in supplemental material (appended to the paper) is recommended, but including URLs to data and code is permitted.
    \end{itemize}

\item {\bf Experimental setting/details}
    \item[] Question: Does the paper specify all the training and test details (e.g., data splits, hyperparameters, how they were chosen, type of optimizer) necessary to understand the results?
    \item[] Answer: \answerYes{}
    \item[] Justification: The paper specifies the task families, prompt structure, public/hidden split procedure, evaluator constraints, metrics, provider/model choices, and four cross-model ablations / analyses (self-repair, example-count, test-time scaling and ensembling, temperature variance) required to interpret the results (Sections~2--4, 6; Appendix~A).
    \item[] Guidelines:
    \begin{itemize}
        \item The answer \answerNA{} means that the paper does not include experiments.
        \item The experimental setting should be presented in the core of the paper to a level of detail that is necessary to appreciate the results and make sense of them.
        \item The full details can be provided either with the code, in appendix, or as supplemental material.
    \end{itemize}

\item {\bf Experiment statistical significance}
    \item[] Question: Does the paper report error bars suitably and correctly defined or other appropriate information about the statistical significance of the experiments?
    \item[] Answer: \answerYes{}
    \item[] Justification: The main results are single full-corpus benchmark runs, but we include a temperature-variance ablation (Section 6.4) reporting population standard deviations across three independent seeds, which provides a direct estimate of run-to-run variability and contextualizes the precision of inter-model comparisons.
    \item[] Guidelines:
    \begin{itemize}
        \item The answer \answerNA{} means that the paper does not include experiments.
        \item The authors should answer \answerYes{} if the results are accompanied by error bars, confidence intervals, or statistical significance tests, at least for the experiments that support the main claims of the paper.
        \item The factors of variability that the error bars are capturing should be clearly stated (for example, train/test split, initialization, random drawing of some parameter, or overall run with given experimental conditions).
        \item The method for calculating the error bars should be explained (closed form formula, call to a library function, bootstrap, etc.)
        \item The assumptions made should be given (e.g., Normally distributed errors).
        \item It should be clear whether the error bar is the standard deviation or the standard error of the mean.
        \item It is OK to report 1-sigma error bars, but one should state it. The authors should preferably report a 2-sigma error bar than state that they have a 96\% CI, if the hypothesis of Normality of errors is not verified.
        \item For asymmetric distributions, the authors should be careful not to show in tables or figures symmetric error bars that would yield results that are out of range (e.g., negative error rates).
        \item If error bars are reported in tables or plots, the authors should explain in the text how they were calculated and reference the corresponding figures or tables in the text.
    \end{itemize}

\item {\bf Experiments compute resources}
    \item[] Question: For each experiment, does the paper provide sufficient information on the computer resources (type of compute workers, memory, time of execution) needed to reproduce the experiments?
    \item[] Answer: \answerYes{}
    \item[] Justification: All crawling, parsing, prompt construction, and local evaluation ran on a consumer Apple-silicon laptop in CPU-only mode (Appendix~\ref{app:repro}); per-provider batch APIs and Together~AI serverless inference were used for model generation with the exact knobs and prices documented (Appendix~\ref{app:model-setup}); per-model batch costs are reported in Appendix~\ref{app:per-model-main}. Total inference spend across the eleven main runs was approximately \$50.79; supplementary ablations added approximately \$3 of further spend.
    \item[] Guidelines:
    \begin{itemize}
        \item The answer \answerNA{} means that the paper does not include experiments.
        \item The paper should indicate the type of compute workers CPU or GPU, internal cluster, or cloud provider, including relevant memory and storage.
        \item The paper should provide the amount of compute required for each of the individual experimental runs as well as estimate the total compute. 
        \item The paper should disclose whether the full research project required more compute than the experiments reported in the paper (e.g., preliminary or failed experiments that didn't make it into the paper). 
    \end{itemize}
    
\item {\bf Code of ethics}
    \item[] Question: Does the research conducted in the paper conform, in every respect, with the NeurIPS Code of Ethics \url{https://neurips.cc/public/EthicsGuidelines}?
    \item[] Answer: \answerYes{}
    \item[] Justification: The study evaluates models on public mathematical content, does not involve human subjects or private data, preserves anonymity in the submission, and discusses release-related limitations and risks (Sections 2 and 7; Appendix~\ref{app:licensing}).
    \item[] Guidelines:
    \begin{itemize}
        \item The answer \answerNA{} means that the authors have not reviewed the NeurIPS Code of Ethics.
        \item If the authors answer \answerNo, they should explain the special circumstances that require a deviation from the Code of Ethics.
        \item The authors should make sure to preserve anonymity (e.g., if there is a special consideration due to laws or regulations in their jurisdiction).
    \end{itemize}

\item {\bf Broader impacts}
    \item[] Question: Does the paper discuss both potential positive societal impacts and negative societal impacts of the work performed?
    \item[] Answer: \answerYes{}
    \item[] Justification: The paper frames the benchmark as a public measurement tool and discusses negative risks including benchmark overfitting, redistribution constraints, and fixed hidden-test leakage in the limitations and release-status discussions (Section 7; Appendix~\ref{app:licensing}).
    \item[] Guidelines:
    \begin{itemize}
        \item The answer \answerNA{} means that there is no societal impact of the work performed.
        \item If the authors answer \answerNA{} or \answerNo, they should explain why their work has no societal impact or why the paper does not address societal impact.
        \item Examples of negative societal impacts include potential malicious or unintended uses (e.g., disinformation, generating fake profiles, surveillance), fairness considerations (e.g., deployment of technologies that could make decisions that unfairly impact specific groups), privacy considerations, and security considerations.
        \item The conference expects that many papers will be foundational research and not tied to particular applications, let alone deployments. However, if there is a direct path to any negative applications, the authors should point it out. For example, it is legitimate to point out that an improvement in the quality of generative models could be used to generate Deepfakes for disinformation. On the other hand, it is not needed to point out that a generic algorithm for optimizing neural networks could enable people to train models that generate Deepfakes faster.
        \item The authors should consider possible harms that could arise when the technology is being used as intended and functioning correctly, harms that could arise when the technology is being used as intended but gives incorrect results, and harms following from (intentional or unintentional) misuse of the technology.
        \item If there are negative societal impacts, the authors could also discuss possible mitigation strategies (e.g., gated release of models, providing defenses in addition to attacks, mechanisms for monitoring misuse, mechanisms to monitor how a system learns from feedback over time, improving the efficiency and accessibility of ML).
    \end{itemize}
    
\item {\bf Safeguards}
    \item[] Question: Does the paper describe safeguards that have been put in place for responsible release of data or models that have a high risk for misuse (e.g., pre-trained language models, image generators, or scraped datasets)?
    \item[] Answer: \answerNA{}
    \item[] Justification: The work does not release model weights or sensitive scraped media; the benchmark consists of public mathematical descriptions and combinatorial objects, so it is not a high-risk release in the sense targeted by this question.
    \item[] Guidelines:
    \begin{itemize}
        \item The answer \answerNA{} means that the paper poses no such risks.
        \item Released models that have a high risk for misuse or dual-use should be released with necessary safeguards to allow for controlled use of the model, for example by requiring that users adhere to usage guidelines or restrictions to access the model or implementing safety filters. 
        \item Datasets that have been scraped from the Internet could pose safety risks. The authors should describe how they avoided releasing unsafe images.
        \item We recognize that providing effective safeguards is challenging, and many papers do not require this, but we encourage authors to take this into account and make a best faith effort.
    \end{itemize}

\item {\bf Licenses for existing assets}
    \item[] Question: Are the creators or original owners of assets (e.g., code, data, models), used in the paper, properly credited and are the license and terms of use explicitly mentioned and properly respected?
    \item[] Answer: \answerYes{}
    \item[] Justification: Asset provenance, derived-benchmark licensing (CC BY 4.0 for data/reports), model-output redistribution status (subject to upstream provider terms), evaluator-and-pipeline licensing (MIT), and release restrictions are documented in a dedicated appendix and accompanying \texttt{LICENSE}; the paper cites FindStat~\citep{findstat} and all evaluated models (\citep{openai_gptoss_2025,qwen3_2025,qwen2_5_2024,liu2024deepseekv3,dubey2024llama3} for open-weight; OpenAI/Anthropic/Google model cards for closed-source) and related benchmark datasets (Appendix~\ref{app:licensing}).
    \item[] Guidelines:
    \begin{itemize}
        \item The answer \answerNA{} means that the paper does not use existing assets.
        \item The authors should cite the original paper that produced the code package or dataset.
        \item The authors should state which version of the asset is used and, if possible, include a URL.
        \item The name of the license (e.g., CC-BY 4.0) should be included for each asset.
        \item For scraped data from a particular source (e.g., website), the copyright and terms of service of that source should be provided.
        \item If assets are released, the license, copyright information, and terms of use in the package should be provided. For popular datasets, \url{paperswithcode.com/datasets} has curated licenses for some datasets. Their licensing guide can help determine the license of a dataset.
        \item For existing datasets that are re-packaged, both the original license and the license of the derived asset (if it has changed) should be provided.
        \item If this information is not available online, the authors are encouraged to reach out to the asset's creators.
    \end{itemize}

\item {\bf New assets}
    \item[] Question: Are new assets introduced in the paper well documented and is the documentation provided alongside the assets?
    \item[] Answer: \answerYes{}
    \item[] Justification: The paper documents the new benchmark, prompt exports, hidden-instance evaluation setup, ablation protocols, and reproducibility artifacts in the main text and appendix (Sections 2--4, 7; Appendix A).
    \item[] Guidelines:
    \begin{itemize}
        \item The answer \answerNA{} means that the paper does not release new assets.
        \item Researchers should communicate the details of the dataset\slash code\slash model as part of their submissions via structured templates. This includes details about training, license, limitations, etc. 
        \item The paper should discuss whether and how consent was obtained from people whose asset is used.
        \item At submission time, remember to anonymize your assets (if applicable). You can either create an anonymized URL or include an anonymized zip file.
    \end{itemize}

\item {\bf Crowdsourcing and research with human subjects}
    \item[] Question: For crowdsourcing experiments and research with human subjects, does the paper include the full text of instructions given to participants and screenshots, if applicable, as well as details about compensation (if any)? 
    \item[] Answer: \answerNA{}
    \item[] Justification: The paper does not involve crowdsourcing or research with human subjects.
    \item[] Guidelines:
    \begin{itemize}
        \item The answer \answerNA{} means that the paper does not involve crowdsourcing nor research with human subjects.
        \item Including this information in the supplemental material is fine, but if the main contribution of the paper involves human subjects, then as much detail as possible should be included in the main paper. 
        \item According to the NeurIPS Code of Ethics, workers involved in data collection, curation, or other labor should be paid at least the minimum wage in the country of the data collector. 
    \end{itemize}

\item {\bf Institutional review board (IRB) approvals or equivalent for research with human subjects}
    \item[] Question: Does the paper describe potential risks incurred by study participants, whether such risks were disclosed to the subjects, and whether Institutional Review Board (IRB) approvals (or an equivalent approval/review based on the requirements of your country or institution) were obtained?
    \item[] Answer: \answerNA{}
    \item[] Justification: The work does not involve human subjects.
    \item[] Guidelines:
    \begin{itemize}
        \item The answer \answerNA{} means that the paper does not involve crowdsourcing nor research with human subjects.
        \item Depending on the country in which research is conducted, IRB approval (or equivalent) may be required for any human subjects research. If you obtained IRB approval, you should clearly state this in the paper. 
        \item We recognize that the procedures for this may vary significantly between institutions and locations, and we expect authors to adhere to the NeurIPS Code of Ethics and the guidelines for their institution. 
        \item For initial submissions, do not include any information that would break anonymity (if applicable), such as the institution conducting the review.
    \end{itemize}

\item {\bf Declaration of LLM usage}
    \item[] Question: Does the paper describe the usage of LLMs if it is an important, original, or non-standard component of the core methods in this research? Note that if the LLM is used only for writing, editing, or formatting purposes and does \emph{not} impact the core methodology, scientific rigor, or originality of the research, declaration is not required.
    %this research? 
    \item[] Answer: \answerYes{}
    \item[] Justification: Evaluating LLMs is the core empirical subject of the paper, and the models, prompting setup, evaluation protocol, and ablation studies are explicitly described in the abstract, model setup, results, and ablation sections (Abstract; Sections 4--7).
    \item[] Guidelines:
    \begin{itemize}
        \item The answer \answerNA{} means that the core method development in this research does not involve LLMs as any important, original, or non-standard components.
        \item Please refer to our LLM policy in the NeurIPS handbook for what should or should not be described.
    \end{itemize}

\end{enumerate}

\end{document}